\documentclass[manuscript,screen, authorversion]{acmart} 

\AtBeginDocument{%
  \providecommand\BibTeX{{%
    \normalfont B\kern-0.5em{\scshape i\kern-0.25em b}\kern-0.8em\TeX}}}

\setcopyright{acmcopyright}
\copyrightyear{2025}
\acmYear{2025}
\acmDOI{XXXXXXX.XXXXXXX}

\acmJournal{CSUR}
\acmVolume{1}
\acmNumber{1}
\acmArticle{1}
\acmMonth{1}

\acmPrice{15.00}
\acmISBN{978-1-4503-XXXX-X/18/06}

\usepackage{tabularx}
\usepackage{babel}
\usepackage{array}
\usepackage{makecell}
\usepackage{multirow}
\usepackage{amsbsy}
\usepackage{amsthm}
\usepackage{amsmath}
\usepackage{graphicx}
\usepackage{rotating}
\usepackage{geometry}
\usepackage[skip=1pt]{caption}
\usepackage{titlesec}
\usepackage{ulem}

\titlespacing\section{0pt}{1pt plus 1.5pt minus 1pt}{1pt plus 1pt minus 1pt}
\titlespacing\subsection{0pt}{1pt plus 1.5pt minus 1pt}{1pt plus 1pt minus 1pt}
\titlespacing\subsubsection{0pt}{1pt plus 1.5pt minus 1pt}{1pt plus 1pt minus 1pt}

\titlespacing*{\subsection}
{0pt} 
{1ex plus 1ex minus .1ex} 
{1ex plus .1ex} 

\newcolumntype{x}[1]{>{\centering\arraybackslash\hspace{0pt}}p{#1}}

\begin{document}

\title{The Evolution of Reinforcement Learning in Quantitative Finance: A Survey}

\author{Nikolaos Pippas}
\email{Nikolaos.Pippas@warwick.ac.uk}

\affiliation{%
  \institution{University of Warwick}
  \city{Coventry}
  \country{United Kingdom}
}


\authornote{\textbf{Conflict-of-Interest \& Disclaimer} — The author is employed by HSBC Asset Management, which also provided partial funding for this research; the views expressed here are solely the author’s and do not necessarily reflect those of HSBC or its affiliates.}

\author{Elliot A. Ludvig}
\email{Elliot.Ludvig@warwick.ac.uk}

\affiliation{%
  \institution{University of Warwick}
  \city{Coventry}
  \country{United Kingdom}
}

\author{Cagatay Turkay}
\email{Cagatay.Turkay@warwick.ac.uk}

\affiliation{%
  \institution{University of Warwick}
  \city{Coventry}
  \country{United Kingdom}
}

\renewcommand{\shortauthors}{Pippas, Ludvig and Turkay}

\begin{abstract}
Reinforcement Learning (RL) has experienced significant advancement over the past decade, prompting a growing interest in applications within finance. This survey critically evaluates 167 publications, exploring diverse RL applications and frameworks in finance. Financial markets, marked by their complexity, multi-agent nature, information asymmetry, and inherent randomness, serve as an intriguing test-bed for RL. Traditional finance offers certain solutions, and RL advances these with a more dynamic approach, incorporating machine learning methods, including transfer learning, meta-learning, and multi-agent solutions. This survey dissects key RL components through the lens of Quantitative Finance. We uncover emerging themes, propose areas for future research, and critique the strengths and weaknesses of existing methods.


\end{abstract}




\maketitle

\section{Introduction\label{sec:Introduction}}

Over the past decade, interest and development in Artificial Intelligence (AI), particularly Reinforcement Learning (RL), have grown significantly. Progress in RL is notable in gaming, with recent algorithms achieving and surpassing human-level proficiency in Go, Chess, and StarCraft \cite{SILVER_2016, SILVER_2018, VINYALS_2019}. These advancements have spurred the exploration of RL in finance, particularly Quantitative Finance (QF). Although RL shows promise, it also faces steep challenges in the complex and dynamic QF domain. These challenges include the unpredictability of financial markets, high computational demands, and the need for improved model interpretability \cite{DOSHI_2017}. In addition, RL applications in finance must address critical challenges such as the transition from simulation to real-world application, sample efficiency, and the balance between online and offline RL settings. This paper reviews RL applications in QF, encompassing key aspects such as Portfolio Management (PM), option pricing, and asset allocation. However, these potential advantages come with the caveat that RL models must be carefully designed and validated to avoid overfitting and ensure robustness in real-world financial markets. 

Prior to RL's introduction to specific domains in finance, traditional Machine Learning (ML) methods were employed in attempts to establish successful trading systems. Atsalakis and Valavanis \cite{ATSALAKIS_2009} provide a comprehensive survey of these practices, including advanced machine learning techniques like Neural Networks (NN) and neuro-fuzzy systems for financial market forecasting. Furthermore, Ozbayoglu et al. \cite{OZBAYOGLU_2020} review recent developments and attempts focused on Deep Learning (DL)\footnote{The term "Deep Learning" can be traced in Dechter \cite{DECHTER_1986}}. The general pattern in these methods consists of a two-step process:
\begin{enumerate}
\item The training of an ML model such as a Support Vector Machine (SVM), NN, or Decision Tree with a specific dataset (features), followed by the generation of a forecast or signal over $n$ periods ahead;
\item The integration of this forecast or signal into a trading system to determine actual trading action or holdings (e.g., buy, sell, or hold in three discrete representations) at a single stock or portfolio level.
\end{enumerate}
Despite the interest in the academic literature, this framework has several limitations \cite{MOODY_1998}, which can potentially be better addressed by RL:
\begin{enumerate}

\item \textbf{Unsuitable optimisation metrics:} In the traditional ML-based approaches, the focus on minimising forecast error is often misaligned with the practical needs of financial trading, where Risk Performance Measures (RPM) like the Sharpe ratio (SR) \cite{SHARPE_1994}\footnote{The Sharpe ratio quantifies the return of an investment exceeding the risk-free rate per unit of risk, as measured by standard deviation.} are more relevant. This disconnect can lead to suboptimal outcomes. RL trading systems, by contrast, naturally optimise for selected RPMs or other desired measures. Nevertheless, a notable exception exists in Mabu et al. \cite{MABU_2007}, which targets optimising forecast error rather than any RPM. Despite its potential to dynamically optimise for various performance measures, RL must contend with the high variability and noise inherent in financial data, which can lead to unpredictable results if not properly managed.


\item \textbf{Limited computational agility:} The two-step process in supervised learning increases complexity and slows predictions. In High-Frequency Trading (HFT) settings, where market conditions change rapidly, these delays can render forecasts quickly irrelevant \cite{GANESH_2018}, highlighting the need for a rapid, online framework that ensures timely decision-making. RL systems, with their ability to learn and adapt in real-time, offer a more agile alternative by continuously updating strategies as new data becomes available, thus better aligning with the fast-paced demands of HFT environments.



\item \textbf{Limited integration of the financial environment:} Traditional methods, such as the conventional portfolio optimisation \cite{MARKOWITZ_1952}, consist of a process that involves two distinct steps: first, calculating the expected returns and the covariance matrix of the assets\footnote{Expected returns can be, for example, the past returns of each asset over a given time window, such as one year.}; second, using these inputs in mean-variance optimisation\footnote{Mean-variance optimisation is a convex optimisation process that balances maximising expected returns and managing risk.} \cite{MARKOWITZ_1952} to manage a predefined risk budget. The separation of these steps can limit the cohesion and adaptability of the framework. In contrast, RL integrates this two-step process into one, promoting a more cohesive framework.

\item \textbf{Limited consideration of constraints:} Traditional frameworks typically incorporate constraints like transaction costs and liquidity in a static manner, relying on predefined assumptions that may not accurately capture the dynamic nature of financial markets. In contrast, RL frameworks allow for real-time integration of these constraints. However, it is important to note that RL models often simplify transaction costs as fixed and assume certainty of execution, overlooking market realities like varying bid/ask spreads \cite{MOODY_1998, BORRAGEIRO_2021, BORRAGEIRO_2022}. Some studies, such as \cite{WANG_2021_B, DENG_2016, FENGQIAN_2020}, have addressed these complexities by incorporating execution and slippage costs.

\item \textbf{Adaptability to Changing Market Conditions:} Financial markets are inherently dynamic and continuously evolving \cite{CONT_2001}. Traditional methods, such as those proposed by Markowitz \cite{MARKOWITZ_1952}, may struggle to capture these changes promptly, often proving slow to adapt to shifting market conditions. 
In contrast, RL algorithms are capable of continuously learning and adapting in real-time, offering an advantage in responding to market shifts, such as regime changes or unexpected events, which traditional models may not effectively manage \cite{MOODY_2001, SUTTON_2018}. Recent work on model-based RL shows that the dynamics model itself can be trained to generalise across unseen transition functions, thereby enabling \textit{zero-shot} adaptation. Notable examples include trajectory-wise multiple-choice learning \cite{SEO_2020}, and context-aware dynamics models \cite{LEE_2020_A}.

\end{enumerate}

Despite its potential, applying RL in QF faces significant challenges. QF's complexity and dynamism considerably surpass RL's conventional learning tasks. In addition, fully representing the financial environment is a near-impossible feat, necessitating substantial effort to create meaningful features. Financial data, laden with noise, exhibit non-stationarity and stochasticity \cite{CONT_2001}, which can lead to unpredictable results. Researchers often counter these challenges by adopting certain simplified assumptions, such as the investor's inability to influence financial markets or the absence of risk aversion, leading to total investment \cite{NEUNEIER_1996, NEUNEIER_1998}. These assumptions imply that the investor or investment is small.  A further challenge lies in maintaining the balance between exploration and exploitation: overemphasising the former could inflate transaction costs. The temporal credit assignment problem \cite{SUTTON_1984}, which arises when the effects of an action are not immediately apparent, is another potential issue in QF \cite{MOODY_2001}. Moreover, interpretability is crucial in finance as stakeholders must understand how an algorithm arrives at its decisions and validate its recommendations. While this 'black box' issue is common across many machine learning models, including deep learning, it is particularly critical in RL due to the sequential decision-making involved. The final notable challenge is the limited availability of historical financial data and the sample inefficiency of RL algorithms, which could restrict the agent's sample space for discovering the optimal policy. In the rest of the paper, we will review several proposed solutions to these challenges, such as the implementation of transfer Learning \cite{PAN_2010}. In section \ref{sec:Results-Discussion}, we also present concerns about evaluation practices, model interpretability, and the potential for overengineering in the literature.

\subsection{Reinforcement Learning and Finance \label{subsec:Reinforcement Learning and Finance}}

\begin{figure}[t!]
\centering
\includegraphics[width=0.4\textwidth]{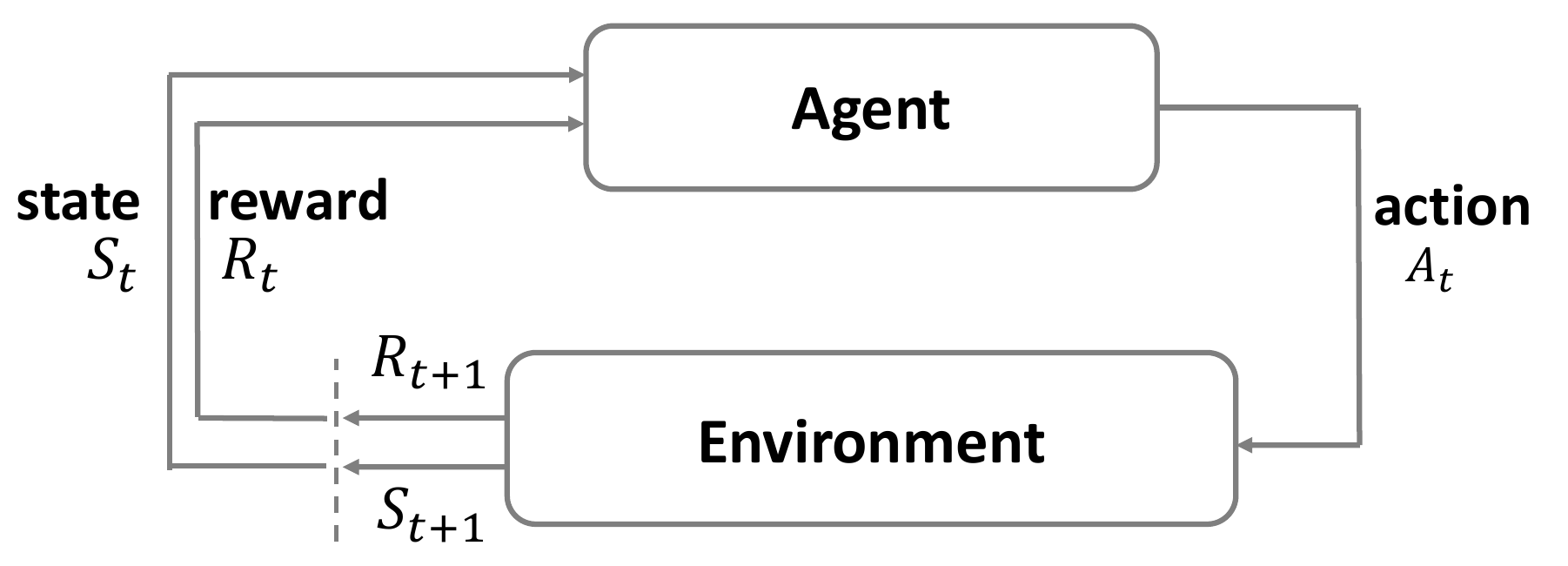}
\caption{The general schematic of agent-environment interaction under the RL framework \cite{SUTTON_2018}.}
\label{fig:Base}
\end{figure}

A good definition of RL is provided by Sutton in \cite{SUTTON_2018}:
``\textit{Reinforcement learning is learning what to do--how to map situations
to actions--so as to maximise a numerical reward signal.}'' This simple sentence conveys all the crucial notions of RL. To unpack this point further, the general flow of an RL system is laid out in Figure \ref{fig:Base}. The agent receives information about the state $S_{t}\in\mathcal{S}$ and a reward at time $t$ and reacts upon this information with actions $A_{t}\in\mathcal{A}$. The resulting action feeds back to the environment, and the system generates a new state $S_{t+1}$ and a new numerical reward $R_{t+1}\in\mathcal{R}\subset\mathbb{R}$ at time $t+1$.\footnote{\(\mathcal{S}\) is the state space, \(\mathcal{A}\) is the action space, and \(\mathcal{R}\) is the reward space (a subset of \(\mathbb{R}\)), representing the possible states, actions, and rewards in reinforcement learning.} The goal is to learn a policy to maximise the expected total reward, which effectively creates a trajectory that can be represented as follows:
\[
S_{0},A_{0},R_{1},S_{1},A_{1},R_{2},...
\]
RL is a framework for solving sequential decision-making
problems. Naturally, many real-world applications fit this framework. For example, video games, robotic control, and driving can be seen
as sequential decision-making problems \cite{MNIH_2013, SILVER_2016, FOX_2024}. The framework above also fits in finance, particularly QF. The agent could be a trader or a portfolio manager who observes the current state of financial markets (the environment) and acts upon this information to maximise a reward function (e.g., SR). Figure~\ref{fig:RL_FLowGraph} illustrates a general RL-based framework in the context of QF, mapping key concepts such as the agent, state, action, reward, and environment. The figure shows several options for these components, depending on the specific QF application context.

\begin{figure}[t!]
\begin{centering}
\includegraphics[scale=0.05]{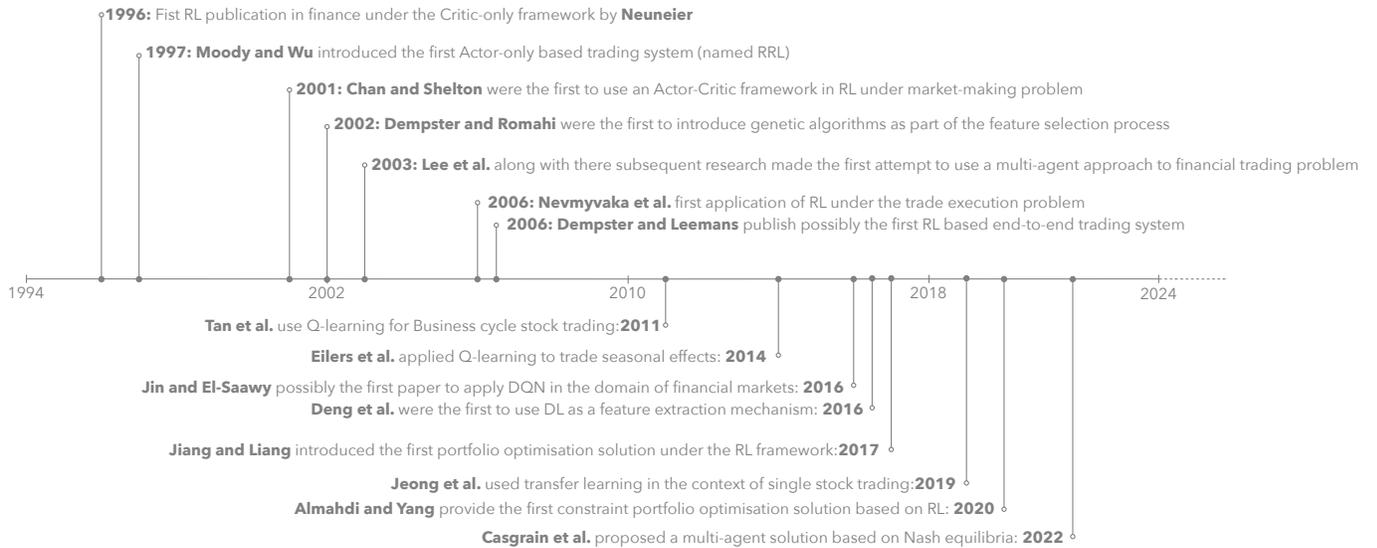}
\par\end{centering}
\caption{\label{fig:timeline} Timeline of important publications in QF under the RL-based framework.}
\end{figure}

Figure \ref{fig:timeline} outlines a timeline of significant RL-based contributions in QF. As is illustrated here, the first explicit applications of RL to finance can be seen as early as the mid-90s with applications of Critic-only (e.g.,~\cite{NEUNEIER_1996}) and Actor-only (e.g.,~\cite{MOODY_1997}) methods. The early 2000s saw the use of Actor-Critic methods (e.g.,~\cite{CHAN_2001}) with deep-learning methods (e.g.,~\cite{JIN_2016}) introduced a bit later with multi-agent solutions (e.g.,~\cite{CASGRAIN_2022}) gaining popularity in recent years.


\subsection{Temporal Dynamics in Financial Applications}

The time unit in the context of QF can vary depending on the specific application. It could be seconds, minutes, hours, days, months, or even years. The choice of time unit depends on the frequency of trading strategies employed. For example, high-frequency trading strategies might use milliseconds or microseconds, while long-term investment strategies might use days, weeks, or, in rare cases, even years. Moreover, in some QF applications, the rewards might not be immediate but delayed. For example, a reward might be given only at $t+12$, representing a delayed return on investment, such as dividends. The RL framework must account for such delayed rewards, ensuring that the agent can still optimise its policy to maximise long-term gains despite the delay. This requires careful consideration of the timing and valuation of future rewards to ensure robust performance.

\subsection{Multi-Agent systems in Finance}

In RL, Multi-Agent Systems (MAS) extend the single-agent paradigm to environments where multiple agents interact, either cooperatively or competitively. The multi-agent model formalism in RL is crucial for addressing complex, dynamic and decentralised decision-making processes found in financial markets \cite{BUSONIU_2008, PLAAT_2022}. Generalising from the single agent, we have:

\begin{itemize}
    \item \textbf{Agents:} Each agent $i$ has its own set of states $S_i$, actions $A_i$, and policies $\pi_i$. Agents can have distinct state spaces $S_i$ depending on their roles and the information they can access. Similarly, agents can have distinct action spaces $A_i$, reflecting their different capabilities or roles in the environment.
    \item \textbf{State Space:} The joint state space $S$ is a combination of all individual states $S = S_1 \times S_2 \times \ldots \times S_n$.
    \item \textbf{Action Space:} The joint action space $A$ is the Cartesian product of all individual action spaces $A = A_1 \times A_2 \times \ldots \times A_n$.
    \item \textbf{Reward Function:} The reward function $R: S \times A \rightarrow \mathbb{R}^n$ provides a vector of rewards for all agents, where each component $R_i$ corresponds to the reward received by agent $i$.
    \item \textbf{Policies:} Each agent follows a policy $\pi_i: S_i \rightarrow A_i$, mapping states to actions.
    \item \textbf{Objective:} The goal is to find a set of policies, $\pi = (\pi_1, \pi_2, \ldots, \pi_n)$, where each policy $\pi_i$ maximises the cumulative reward for its respective agent. Depending on the application, this may involve maximising the cumulative reward for all agents collectively or maximising individual rewards independently.

\end{itemize}

Interaction is crucial for modelling realistic financial systems in which traders might share market information or collaborate on trading strategies. In multi-agent RL (MARL), agents can interact in various ways, including:

\begin{itemize}
\item \textbf{Information Sharing:} Agents share observations or information to enhance decision-making, leading to more informed actions and better performance \cite{TAN_1993}.
\item \textbf{Shared States:} Agents access a global or subset of shared states, enabling coordinated actions \cite{FOERSTER_2017}.
\item \textbf{Joint Actions:} Agents coordinate actions to achieve common goals, such as synchronising trades to influence market prices \cite{GUESTRIN_2001}.
\item \textbf{Cooperative and Competitive Interactions:} Agents work cooperatively for joint rewards or competitively for individual rewards, depending on the financial application \cite{BUSONIU_2010}.
\end{itemize}

In the context of finance, a multi-agent RL setup might involve various trading agents operating simultaneously, each with its distinct strategy and goal. For example, one agent might focus on long-term investments, while another specialises in high-frequency trading. These agents interact with each other and the market, influencing and responding to market dynamics. By incorporating multi-agent models, we can better simulate financial markets' complex and competitive nature. Finally, we discuss applications of MAS solutions in QF in subsection \ref{subsec:Multi-agent-applications}.
\begin{figure}[t]
\centering
\includegraphics[scale=0.5]{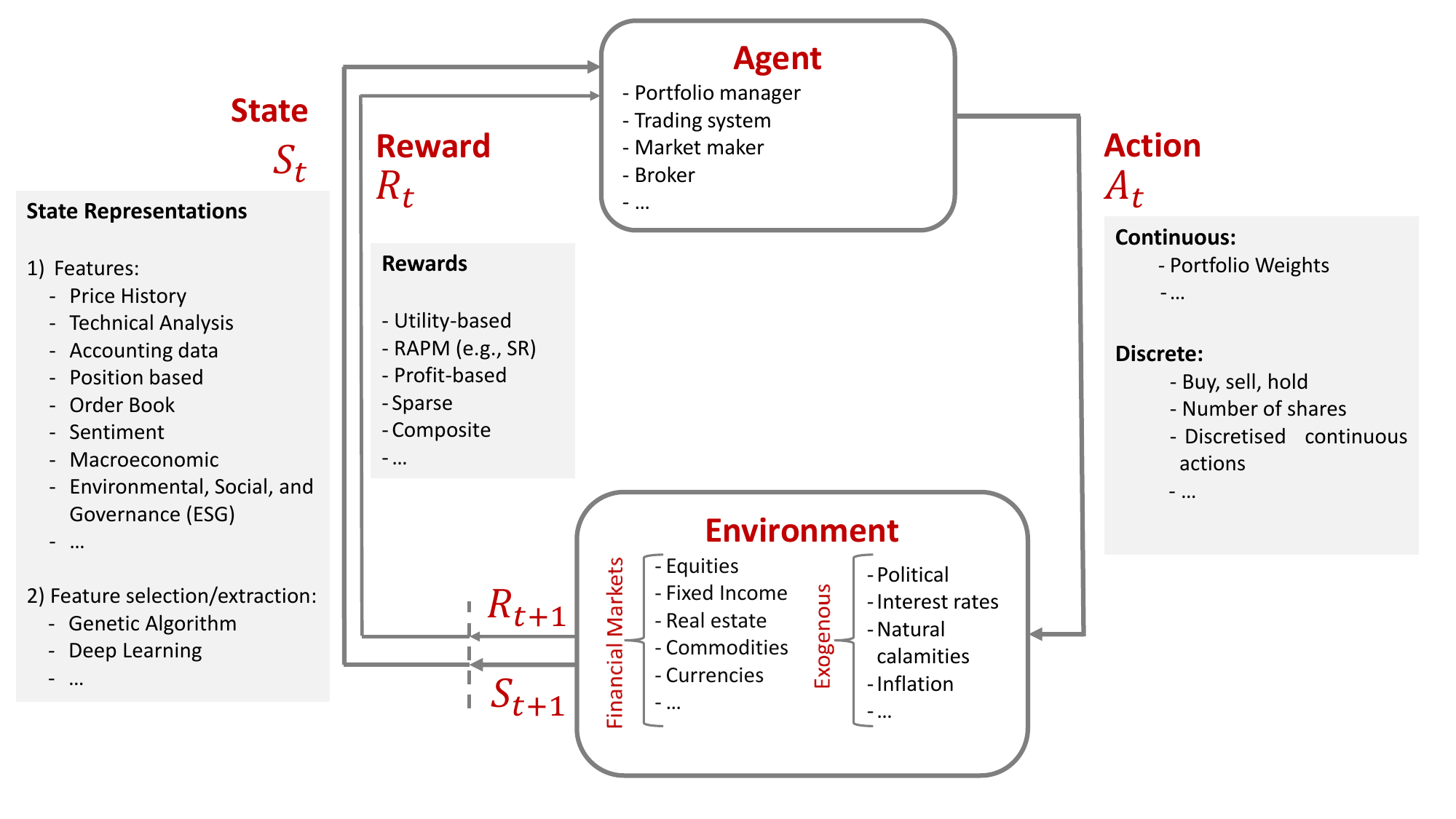}
\caption{The overview specifies how an agent and the environment interact in the QF domain using the classical RL framework depicted in Figure~\ref{fig:Base}. With this, we map concepts, techniques and practices from QF that are identified in the survey to the components of the RL framework. (Note: Profit-based rewards can include financial gains such as dividends and pay-offs)}
\label{fig:RL_FLowGraph}
\end{figure}
\subsection{Contribution and Paper Organisation}

This survey assesses 167 publications to provide the first comprehensive review of the critical components of an RL agent within QF. Furthermore, our review demonstrates the diverse applicability of RL in different QF contexts. Although the recent surge in interest in RL within finance has produced several survey publications \cite{FISCHER_2018, MOSAVI_2020, CHARPENTIER_2021, HAMBLY_2023}, our study is distinctively focused on QF, diverging from other studies that encompass broader economic applications \cite{MOSAVI_2020, CHARPENTIER_2021} and rooted more closely within the literature on finance theory and practice. Moreover, we avoid comprehensive textbook reviews of RL theory such as those presented in other surveys \cite{HAMBLY_2023}, opting for the necessary definitions and references as necessary.

The articles for this survey were sourced from Google Scholar using RL and finance keywords and employing the snowball method \cite{WOHLIN_2014} on relevant references, spanning 1996-2022. In particular, only 27.5 \% of the publications predate 2016, reflecting the recent surge in interest and contribution of RL to QF. This can be mainly attributed to the introduction of Deep Q Networks (DQN) \cite{MNIH_2013, MNIH_2015}, and consequently, Deep Reinforcement Learning (DRL).

In other research fields, related concepts were known as trial-and-error learning, learning with a critic, optimal control, and dynamic programming. Therefore, a range of  papers~\cite{GARLEANU_2013, JOHANNES_2014, HUO_2017} link their research to concepts similar to those in RL. Additionally, while multi-armed bandits (MABs) and sequential optimisation methods are significant in online learning and decision-making, they differ from the broader RL framework covered in this survey. MAB problems are typically simpler and do not involve state transitions. Thus, this survey focuses on comprehensive RL frameworks in QF while acknowledging key MAB research, such as \cite{HUTTER_2005, AUDIBERT_2009, AUER_2002}. Thus, we focus solely on methods that explicitly fall into the RL research area in QF.

This survey starts with dissecting common RL methods and key RL components, e.g., the agent, environment and rewards, through the lens of Quantitative Finance. Building on an in-depth dialogue with quantitative finance literature and practice and reflecting on recent advances in machine learning, we uncover emerging themes, propose areas for future research, and critique the strengths and weaknesses of existing methods.
\section{Critical Considerations for RL in QF\label{sec:Critcal_RL_Dimension_in_QF}}

\noindent Before delving into the research and implementation of RL in QF, we introduce some critical dimensions and considerations of RL in QF. Figure~\ref{fig:RL_FLowGraph}, as we covered in subsection  \ref{subsec:Reinforcement Learning and Finance}, expands the generic RL framework and illustrates several options for potential mappings between key RL components and relevant concepts, techniques and practices from QF. In addition to these, there are some critical considerations for incorporating RL in QF including the transition from simulation to real-world applications, sample efficiency, online versus offline RL settings, on-policy versus off-policy frameworks, and how RL interacts with the unique challenges of financial markets.
\subsection{Transition from Simulation to Real-World Application\label{subsec:Tran_Real_World}}

Deploying RL models in real financial markets presents unique challenges. Unlike controlled simulation environments, real-world markets are unpredictable and influenced by myriad factors such as information asymmetry, variable transaction costs, taxes, and noise traders, which can significantly affect model performance \cite{STIGLITZ_1981, MOODY_1998}. Therefore, a robust validation and simulation process is essential to ensure that RL models accurately capture the complexities of chaotic financial markets. Incorporating domain knowledge improves model resilience and adaptability, providing a more accurate reflection of market conditions, and improving generalisation capabilities. Furthermore, developing a comprehensive risk management framework \cite{MIHATSCH_2002, DEMPSTER_2006} is vital to mitigate regulatory and stakeholder risks. This framework should include thorough backtesting, stress testing, and continuous monitoring to ensure that RL models can adapt to market changes and maintain their performance in various scenarios.

\subsection{Sample Efficiency\label{subsec:Sample_Efficiency}}

Improving sample efficiency is crucial in RL, especially in financial applications where data can be sparse, have a short history, or be expensive to obtain \cite{HAMBLY_2023}. Many samples drawn from the environment are often necessary for stable convergence and low variance. Techniques such as model-based RL, experience replay, and transfer learning are valuable in this context.

Model-free methods are typically slower to learn, whereas model-based methods build a transition model from the environment's feedback, allowing the agent to use this model to understand the effects of actions on states and rewards \cite{PLAAT_2022}. Model-based RL can generate additional training data, enhancing learning efficiency \cite{SILVER_2016}. For instance, model-based RL can simulate various market conditions to create synthetic data in a stock-trading scenario. This data helps the model learn how to handle different market situations. However, for this approach to be effective, the model must accurately represent the environment. Experience replay allows RL agents to reuse past experiences, thereby improving learning from limited data \cite{MNIH_2015}. In the context of financial applications, agents can reuse previous experiences to improve learning efficiency. Transfer learning enables leveraging knowledge from related tasks to improve performance in new but similar tasks, thus addressing data scarcity \cite{LAZARIC_2012}. For a discussion of transfer learning and model-based RL, see subsection \ref{subsec:Transfer-learning} and Section \ref{sec:Future-Aseas-for-Research}. When the transition model is accurate, model-based RL can also serve as a transfer learning mechanism, further enhancing learning efficiency \cite{PLAAT_2022}.

\subsection{Online vs. Offline RL Settings\label{subsec:Online vs Offline}}

Online learning in QF refers to agents learning and adapting continuously as new data arrive. This approach benefits from the ability to dynamically update strategies in response to market changes. However, this can be highly impractical in HFT, where decisions occur in microseconds to milliseconds, because online learning requires real-time updating of algorithm parameters and significant computational resources to process new data. As pointed out by Hambly et al.~\cite{HAMBLY_2023}, a potential solution in this setup might be to collect data with a pre-specified exploration scheme during trading hours and update the algorithm with newly collected data after trading closes. Transfer learning can also be leveraged in this context to bridge the gap between offline and online learning. Initially, the RL agent can be pre-trained using offline RL on historical data, learning general strategies and patterns. Once deployed, the agent can then use online learning to fine-tune these strategies in response to real-time market conditions. This hybrid approach can mitigate some of the computational burdens of online learning by reducing the frequency and extent required for real-time updates. Also, certain components of the RL framework can be constructed to support online learning, such as the introduction of Differential SR \cite{MOODY_1997, MOODY_1998, MOODY_1998_2, MOODY_2001}, which is an online version of SR (see subsection \ref{subsec:Rewards}). 

In contrast, offline RL uses historical data to develop strategies without continuous interaction with the environment. This method is particularly useful for backtesting and strategy development, where real-time data streams are not necessary. Offline RL can leverage vast amounts of historical market data to train models and then be deployed in live environments after thorough validation. Furthermore, offline learning does not face the computational constraints that online learning faces, allowing for extensive preprocessing and model training using conventional computing resources. Though training time can be extensive due to large datasets, it avoids the real-time computational demands of online RL.

\subsection{On-Policy vs. Off-Policy Frameworks\label{subsec:ONvsOFF Policy}}

The distinction between on-policy and off-policy methods significantly affects their application in various financial contexts, each presenting unique challenges and opportunities.

On-policy algorithms, such as Proximal Policy Optimisation (PPO) \cite{SCHULMAN_2017}, use data generated only from the current policy. This means that only actions taken by the current policy generate data at each training iteration. Consequently, the data become unusable after policy updates, making on-policy methods inefficient. However, on-policy algorithms are typically more stable and exhibit lower variance because they learn directly from the policy they are improving \cite{PLAAT_2022}. For example, SARSA \cite{RUMMERY_1994}, an on-policy algorithm, is used in stock trading models due to its ability to adapt to market changes efficiently, resulting in lower computational resources requirements and robust performance under varying market conditions \cite{CORAZZA_2015, DEOLIVEIRA_2020}. Furthermore, on-policy methods could be practical in HFT due to the vast amount of data available despite being sample inefficient. In PM, trading at lower frequencies (e.g. monthly) could be less practical due to the scarcity of available data.

In contrast, off-policy methods, such as DQN \cite{MNIH_2015}, do not have this limitation, allowing them to use data generated from different policies. This makes off-policy methods more sample-efficient, as they can reuse past experiences stored in memory. However, this requires significant memory storage to preserve past experiences. Additionally, off-policy methods can face challenges with stability and variance due to the complexity of updating policies from a mixture of different data sources and policies, which can amplify approximation errors and lead to instability \cite{PLAAT_2022, SUTTON_2018}. Off-policy algorithms often leverage large datasets for training and can learn from historical data, making them particularly suitable for environments where collecting real-time data is impractical. For example, the application of Q-learning \cite{WATKINS_1989} in financial trading demonstrates its effectiveness in using past market data to inform future trading decisions, which could lead to better adaptation to non-stationary market dynamics \cite{CORAZZA_2015}. Furthermore, off-policy methods could be used in HFT due to their ability to process large amounts of data, although they may require longer training times, which is crucial in the HFT context. In PM, trading at lower frequencies (e.g. monthly) could benefit more from off-policy methods due to their better use of available data.


\section{Main Reinforcement Learning Methods \label{sec:Main-Reinforcement-Leaning}}

Contemporary literature distinguishes four primary RL methods: Value-based (Critic-only), Policy-based (Actor-Only), Actor-Critic, and model-based. This section delves into a thorough exploration of the principal RL techniques encapsulated within these categories. We provide an overview of the general framework, the advantages and disadvantages, and the applicability of each method within the field of QF. Additionally, within these categories, we incorporate other forms of RL approaches such as multi-agent methods \cite{LEE_2002, HUANG_2022}. 

Table \ref{tab:RL Method} presents a classification of each publication reviewed in this survey according to the RL method and algorithm used. Within the Value-based approach, DQN and Q-learning \cite{WATKINS_1989} emerge as the most frequently used algorithms. Recurrent Reinforcement Learning (RRL) \cite{MOODY_1997, MOODY_1998, MOODY_1998_2, MOODY_2001} assumes dominance in the Policy-based methodology, and PPO \cite{SCHULMAN_2017}, and DDPG \cite{LILLICRAP_2015} are the most popular in the Actor-Critic approach. Overall, the Value-based method has received the most extensive research, followed by the Policy-based, Actor-Critic, and model-based methods. In the interest of simplicity, we have occasionally categorised publications under the ``other'' bucket when the proposed algorithm is non-standard or infrequently used. Furthermore, we have classified several publications under multiple categories due to their use of a diverse range of algorithms and refrained from including publications that fall outside these four categories. 

\begin{table}[htbp]
\centering
\begin{tabular}{|c|c|c|l|c|}
\hline
\textbf{Approach} & \textbf{Method} & \textbf{RL-Algorithm} & \textbf{Publication} & \textbf{Count} \\
\hline 
\multirow{19}{*}{Model-Free} & \multirow{9}{*}{\makecell{Value-Based \\ (Critic-Only)}} & \multirow{4}{*}{DQN} & \cite{ABDELKAWY_2021}, \cite{CANNELLI_2020}, \cite{CARAPUCO_2018},
\cite{CARTA_2021}, \cite{CARTEA_2021}, \cite{DABERIUS_2019}, \cite{DU_2020}, \cite{HUANG_2018},
\cite{HUANG_2022}, \cite{JEONG_2019}, \cite{KARPE_2020}, \cite{KIM_2019},  & \multirow{4}{*}{37} \\
\cline{4-4} 
 &  &  & \cite{KIM_2022}, \cite{KUMAR_2022}, \cite{LEE_2020}, \cite{LEEM_2020}, \cite{LI_2020},
\cite{LU_2022}, \cite{LUCARELLI_2019}, \cite{NAN_2020}, \cite{NING_2020},
\cite{PARK_2020}, \cite{PATEL_2018},  & \\
\cline{4-4} 
 &  &  & \cite{SHAVANDI_2022}, \cite{SORMAYURA_2019}, \cite{TAGHIAN_2022}, \cite{THEATE_2021},
\cite{TSANTEKIDIS_2020}, \cite{WANG_2017}, \cite{WANG_2021_B},
\cite{WANG_C_2021}, \cite{WEI_2019}, \cite{WU_2020}, \cite{YUAN_2020},  & \\
\cline{4-4} 
 &  &  & \cite{ZARKIAS_2019}, \cite{ZHANG_2020_A}, \cite{ZHAO_2021} & \\
\cline{3-5}
 &  & \multirow{4}{*}{Q-Learning} & \cite{BATES_2003}, \cite{BERTOLUZZO_2012}, \cite{CAO_2021}, \cite{CHAKOLE_2021},
\cite{CORAZZA_2014}, \cite{CORAZZA_2015}, \cite{DEMPSTER_2001,DEMPSTER_2002},
\cite{DU_2009}, \cite{DUBROV_2015}, \cite{EILERS_2014}, \cite{FEUERRIEGEL_2016}, \cite{GAO_2000} & \multirow{4}{*}{36} \\
\cline{4-4} 
 &  &  & \cite{HALPERIN_2020}, \cite{HENDRICKS_2014}, \cite{HRYSHKO_2004},
\cite{JANGMIN_2006}, \cite{JIN_2016}, \cite{KAUR_2017}, \cite{LEE_2002,LEE_2002_2,LEE_2003,LEE_2007},
\cite{LI_2006}& \\
\cline{4-4} 
 &  &  & \cite{LI_2009}, \cite{LIM_2018}, \cite{LU_2022}, \cite{NEUNEIER_1996,NEUNEIER_1998},
\cite{NEVMYVAKA_2006},  \cite{PENDHARKAR_2018}, \cite{SHEN_2014}, \cite{SPOONER_2018},
\cite{TAN_2011}, \cite{ZHONG_2021}, & \\
\cline{4-4}
 &  &  & \cite{ZHU_2018} & \\
\cline{3-5} 
 &  & SARSA & \cite{CHAN_2001}, \cite{CHEN_2007}, \cite{CORAZZA_2015}, \cite{DEOLIVEIRA_2020},
\cite{GU_2011}, \cite{KOLM_2020}, \cite{PENDHARKAR_2018}, \cite{SHERSTOV_2004},
\cite{SPOONER_2018}, \cite{TAGHIAN_2022} & 10 \\
\cline{3-5} 
 &  & Other & \cite{ALAMEER_2021}, \cite{CANNELLI_2020}, \cite{DIXON_2020}, \cite{DUBROV_2015}, \cite{LI_2009},\cite{PENDHARKAR_2018} & 6 \\
\cline{2-5}
 & \multirow{6}{*}{\makecell{Policy-Based \\ (Actor-Only)}} & \multirow{2}{*}{PG} & \cite{BENHAMOU_2020_A,BENHAMOU_2020_B,BENHAMOU_2021_A,BENHAMOU_2021_B},
\cite{BUEHLER_2019}, \cite{CONG_2020}, \cite{JIA_2019}, \cite{LIANG_2018}, \cite{WANG_2019},
\cite{WANG_2021_A}, \cite{ZHANG_2020_A}, 
& \multirow{2}{*}{13} \\
\cline{4-4} 
& &  & \cite{ZHANG_2020_B}, \cite{ZHAO_2021} & \\
\cline{3-5} 
 &  & REINFORCE & \cite{CORQUERET_2022}, \cite{DING_2018}, \cite{DU_2009}, \cite{WANG_2021_B}, \cite{WEI_2019} & 5 \\
\cline{3-5} 
 &  & \multirow{3}{*}{RRL} & \cite{ABOUSSALAH_2020}, \cite{ALMAHDI_2017,ALMAHDI_2019}, \cite{BERTOLUZZO_2007}, \cite{BISHT_2020}, \cite{BORRAGEIRO_2021,BORRAGEIRO_2022}, \cite{DEMPSTER_2006}, \cite{DENG_2015,DENG_2016},
\cite{GABRIELSSON_2015}, \cite{GAO_2000}, \cite{GOLD_2003}, \cite{GORSE_2011}, & \multirow{3}{*}{31} \\
\cline{4-4} 
 &  &  & \cite{GORSE_2011}, \cite{HENS_2007}, \cite{LI_2007}, \cite{LI_2021}, \cite{LU_2017},
\cite{MARINGER_2010,MARINGER_2012,MARINGER_2014}, \cite{MOODY_1997,MOODY_1998,MOODY_1998_2,MOODY_2001},  & \\
\cline{4-4} 
 &  &  & \cite{SHI_2019}, \cite{SI_2017}, \cite{XU_2021}, \cite{ZHANG_2013,ZHANG_2014,ZHANG_2016} & \\
\cline{3-5} 
 &  & Other & \cite{FENGQIAN_2020}, \cite{LEI_2020}, \cite{QUI_2021}, \cite{SATTAROV},
\cite{WANG_2019_B}, \cite{WANG_2020}, \cite{WENG_2020} & 7 \\
\hline 
\multirow{7}{*}{Model-Free} & \multirow{6}{*}{Actor-Critic} & A2C & \cite{KATONGO_2021}, \cite{WEI_2019}, \cite{YANG_2020},  \cite{ZHANG_2020_A} & 4 \\
\cline{3-5} 
 &  & DPG & \cite{ABOUSSALAH_2022}, \cite{JIANG_2017,JIANG_2017_1}, \cite{LIU_2020}, \cite{WU_2020},
\cite{YE_2020} & 6 \\
\cline{3-5} 
 &  & \multirow{2}{*}{DDPG} & \cite{ABDELKAWY_2021}, \cite{ABOUSSALAH_2022}, \cite{BAO_2019}, \cite{CAO_2021}, \cite{KATONGO_2021}, \cite{KORATAMADDI_2021},
\cite{LI_2019}, \cite{LIANG_2018}, \cite{SAWHNEY_2021}, \cite{WANG_2021_D}, \cite{XIONG_2018},
  & \multirow{2}{*}{14} \\
\cline{4-4} 
 &  &  & \cite{YANG_2020}, \cite{YU_2019}, \cite{YU_2020}, & \\
\cline{3-5} 
 &  & \multirow{2}{*}{PPO} &  \cite{ABOUSSALAH_2022}, \cite{BRIOLA_2021}, \cite{DABERIUS_2019}, \cite{DU_2020}, \cite{FANG_2021}, \cite{GANESH_2019}, 
\cite{HUANG_2022}, \cite{KATONGO_2021}, \cite{LIANG_2018}, \cite{LIN_2020},
  & \multirow{2}{*}{14} \\
\cline{4-4} 
 &  &  & \cite{TSANTEKIDIS_2020,TSANTEKIDIS_2021}, \cite{YANG_2020}, \cite{YUAN_2020} & \\
\cline{3-5} 
 &  & TRPO & \cite{BISI_2019}, \cite{VITTORI_2020} & 2 \\
\cline{3-5} 
 &  & Other &  \cite{ABOUSSALAH_2022}, \cite{BEKIROS_2010}, \cite{CASGRAIN_2022}, \cite{CHAN_2001}, \cite{GARCIA_2019}, \cite{GUEANT_2019},
\cite{KIM_2022}, \cite{LI_2007}, \cite{MABU_2007}, \cite{PONOMAREV_2019},
\cite{SOLEYMANI_2021}, \cite{YUAN_2020} & 12 \\
\hline 
\multirow{1}{*}{Model-Based} & \multirow{1}{*}{Model-Based} & Other & \cite{YU_2019}, \cite{GUEANT_2019}, \cite{WEI_2019} & 3 \\
\hline
\end{tabular}
\caption{Categorisation of each publication based on RL methods and algorithms.}
\label{tab:RL Method}
\end{table}

\subsection{Value-Based Methods\label{subsec:Critic-Only-method}}

\subsubsection{\textbf{Core Framework for Value-based RL in QF.}}

Value-based methods are foundational in RL. The agent's objective is to learn the value of different actions in various states to maximise cumulative rewards. Neuneier's work \cite{NEUNEIER_1996} is one of the earliest contributions in this field, which set a methodological precedent that has informed many subsequent studies. This section outlines a generic framework derived from these methodologies:

\begin{enumerate}
    \item \textbf{Define a Finite Set of States \( S_t \):}
    States represent environment-derived information at each time point \( t \in \{1, 2, \ldots, T\} \). This information includes financial accounting data, prices, sentiment, and technical indicators.
    
    \item \textbf{Define a Set of Actions \( A_t \in \{Buy, Sell, Hold\} \):}
    These are the possible actions of the agent at each time \( t \in \{1, 2, \ldots, T\} \).
    
    \item \textbf{Establish Transition Probabilities:}
    These probabilities, typically unmodelled, define state transitions based on actions.
    
    \item \textbf{Formulate a Reward Function \( R_t \):}
    Provides numerical feedback to the agent in response to its preceding action.
    
    \item \textbf{Create a Policy \( \pi \):}
    Maps states to actions for the agent to follow.
    
    \item \textbf{Construct a Value Function \( V \):}
    Maps states to the agent's expected total discounted reward from a given state until the episode's end under policy \( \pi \).
\end{enumerate}

The agent continuously interacts with the market by processing market data, making investments, and receiving returns over many trials. Through this process, the agent aims to discover the best possible strategy, known as the optimal policy ($\pi=\pi^{*}$). This entire approach is an example of a Markov Decision Problem (MDP) applied to financial trading. This framework highlights the essential components required for Value-based RL in finance. By defining states, actions, transition probabilities, reward functions, policies, and value functions, we can create a system where an agent learns to make optimal trading decisions through continuous interaction with the market. This framework, despite its simplicity, serves as a building block for more advanced frameworks.

\subsubsection{\textbf{General Observations, Comments and Definitions for Value-based Methods.}}

Value-based methods are a well-established branch of research in RL applied to trading systems, despite their noted shortcomings. A significant challenge is the discrete nature of the action space, complicating practical trading applications \ref{subsec:Action}.


The agent creates a value function to estimate the results of actions such as buying, holding, or selling, helping to choose the best action. This approach often uses model-free RL algorithms like Q-learning \cite{WATKINS_1989} and SARSA \cite{RUMMERY_1994} to optimise the expected total reward. Q-learning, an off-policy RL algorithm, is proven to converge to the optimal solution in the tabular setting under specific conditions \cite{WATKINS_1992}. In Q-learning, the Q-values, $Q\left(S_{t}, A_{t}\right)$, represent the expected rewards for taking an action $A_{t}$ in state $S_{t}$. These Q-values are updated according to the rule:
\begin{equation}
Q\left(S_{t}, A_{t}\right) \leftarrow Q\left(S_{t}, A_{t}\right) + \alpha \left[R_{t+1} + \gamma \underset{a}{\max} Q\left(S_{t+1}, a\right) - Q\left(S_{t}, A_{t}\right)\right].
\label{eq:-21}
\end{equation}
This update rule adjusts the current Q-value by incorporating the observed reward $R_{t+1}$ and the maximum estimated future reward, weighted by the learning rate $\alpha \in \left(0,1\right]$. Over time, this process helps refine the Q-values, guiding the agent toward the optimal action values under specific conditions. Specifically, Q-learning is guaranteed to converge to the optimal solution with probability 1, provided that all state-action pairs are visited infinitely many times and the learning rate satisfies certain decay conditions. These convergence conditions are crucial for ensuring that the algorithm performs well in practice.


Q-learning and SARSA are traditionally used in tabular settings, which are suitable for small, discrete state spaces. However, these algorithms can also be adapted to more complex state spaces through function approximation techniques. While tabular methods are inherently limited in scalability, FA allows these algorithms to generalize from similar states, making them applicable in real-world scenarios with large or continuous state spaces \cite{SUTTON_2018}. Despite this adaptability, using function approximation introduces practical challenges, such as the risk of overfitting and increased computational demands.

The field experienced a transformative breakthrough with the introduction of DQNs by Mnih et al. \cite{MNIH_2013, MNIH_2015}, leveraging neural networks as function approximators to achieve human-level performance in Atari video games. This amalgamation of RL with deep learning is known as Deep Reinforcement Learning (DRL), and in Q-learning's specific context, it is known as DQL or DQN. The authors used experience replay \cite{LIN_1992} and target networks for a stable DRL agent. Subsequent proposed improvements have included a Prioritised Replay Buffer \cite{SCHAUL_2015}, Double Q-learning (DDQN) \cite{vanHASSELT_2016-1}, Multi-step learning \cite{SUTTON_1988}, Noisy Networks \cite{FORTUNATO_2017}, Dueling DQN \cite{WANG_2016}, and Distributional RL \cite{BELLEMARE_2017}.

In conclusion, the contemporary surge of RL in QF can be traced back to the advent of DQN. Despite the constraints of the Value-based framework, it remains a vibrant research area with numerous significant publications within this survey's scope. The subsequent section explores Policy-based methods that address some of these limitations by directly optimising the policy.

\subsection{Policy-Based Method\label{subsec:Actor-Only-method}}

\subsubsection{\textbf{Core Framework for Policy-Based RL in QF}.}

Policy-based methods in RL focus on directly optimising the policy that dictates the agent’s actions. Unlike Value-based methods, which estimate the value of actions to derive the best policy, Policy-based methods directly search for the optimal policy that maximises the cumulative reward. This approach can be particularly advantageous in environments with continuous action spaces and complex dynamics.

Among these methods, the RRL framework stands out due to its extensive use in the reviewed literature and its clear relevance to finance \cite{MOODY_1997, MOODY_1998, MOODY_1998_2, MOODY_2001}. The RRL framework is particularly suited for financial applications because it can capture the temporal dependencies and sequential nature of trading decisions. Therefore, this subsection will focus on the RRL framework. Following Moody and Wu \cite{MOODY_1997}, an agent’s actions are represented by:
\begin{equation}
A_{t} = A\left(\theta_{t} : A_{t-1}, I_{t}\right) \in \left\{ -1, 0, 1 \right\} \text{ with } I_{t} = \left\{ z_{t}, z_{t-1}, z_{t-2}, \ldots, y_{t}, y_{t-1}, y_{t-2}, \ldots \right\}.
\label{eq:-3}
\end{equation} Here, $\theta_{t}$ refers to the learned parameters of the agent, $A_{t}$ corresponds to a trading position that can assume one of three states - sell, neutral, or buy, while $A_{t-1}$ represents the preceding action at time $t-1$, $I_{t}$, which denotes the information set at time $t$, is used as a state representation, composed of lagged asset prices $z_{t}$ and other external variables $y_{t}$. $A_{t}$ can be defined \cite{MOODY_2001} as:
\begin{equation}
A_{t}=sign\left(uA_{t-1}+v_{0}r_{t}+v_{1}r_{t-1}+...+v_{m}r_{t-m}+\omega\right),\label{eq:-4}
\end{equation}
where, $r_{t}$ denotes the price return, and $\theta_{t} = \theta = \{u, v_{i}, w\}$ with \( i \in \{0, 1, \ldots, m\} \). The RRL framework further optimises the trading system by maximising a performance function $U_{t}$. This function can represent different goals, such as maximising profit, improving a utility function of wealth, or enhancing performance ratios like the Sharpe Ratio (SR). Essentially, it helps the trader aim for better financial performance based on their specific goals. One of those choices is the additive profits utility reward function \cite{MOODY_2001}:
\begin{equation}
U_{t}(\theta)=P_{T}=\sum_{t=1}^{T}R_{t}=\mu\sum_{t=1}^{T}\left\{ r_{t}^{f}+A_{t-1}\left(r_{t}-r_{t}^{f}\right)-\delta_{t}\left|A_{t}-A_{t-1}\right|\right\} ,\label{eq:-5}
\end{equation}
where, $P_{T}$ denotes the cumulative profit at the end of trading period $T$, $R_{t}$ the profit or loss at time $t$, and $T$ the total time steps. $\mu>0$ represents a fixed position size when buying/selling a stock, $r_{t}$ and $r_{t}^{f}$ are the absolute prices changes for the period from $t-1$ to $t$ for the risky asset (stock) and the risk-free asset (e.g., T-Bills), respectively, and $\delta_{t}$ denotes transaction costs from buying/selling the risky asset. 

An important aspect of the RRL framework is the online optimisation approach using stochastic gradient ascent to adjust the parameters. This process involves calculating how changes in the agent's strategy (represented by the parameters $\theta$) affect its performance (measured by the utility function $U_{t}$). Essentially, it looks at how both the current and previous actions influence the agent's success, allowing for continuous improvement. Imagine a trader adjusting their strategy based on both their recent trades and past experiences to maximise profits. The gradient calculation is described by the following equation:
\begin{equation}
\frac{dU_{t}\left(\theta\right)}{d\theta}=\frac{dU_{t}\left(\theta\right)}{dR_{t}}\left\{ \frac{dR_{t}}{dA_{t}}\frac{dA_{t}}{d\theta}+\frac{dR_{t}}{dA_{t-1}}\frac{dA_{t-1}}{d\theta}\right\}. \label{eq:-6}
\end{equation}
This equation shows that we need to compute the gradient of the utility function with respect to the actions and then update the parameters based on this gradient. The update is done using the learning rate $\rho$, as in $\Delta\theta_{t}=\rho\frac{dU_{t}\left(\theta\right)}{d\theta_{t}}$. In simpler terms, we adjust our model parameters by considering how both the current and previous actions influence the reward, and we use this information to make our model more accurate over time.

Numerous versions of this RRL framework have been proposed. For instance, the study by Gorse et al. \cite{GORSE_2011} presents a learning rule based on associative reward-penalty \cite{BARTO_1985}, with the output from equation (\ref{eq:-4}) defined as a probability. Gold \cite{GOLD_2003}, inspired by \cite{MOODY_1997}, introduced an additional hidden layer in the network, capturing more complex patterns compared to the single-layer network, thus serving as a precursor to Deep Recurrent Reinforcement Learning (DRRL) \cite{DENG_2016}.

\subsubsection{\textbf{General Observations, Comments and Definitions for Policy-Based Methods}.}

Policy-based methods represent the second most explored area of RL in the reviewed literature. These methods provide a direct mapping from states to actions, eliminating the need to compute the expected outcome of different actions as in the Value-based approach, resulting in faster learning processes. A significant advantage of Policy-based methods is the continuous action space for the agent. Consider a portfolio of stocks: with the Value-based approach, portfolio weights can only take discrete values like buy, sell, or hold. In contrast, the Policy-based approach allows portfolio weights to assume any value in $\left[0,1\right]$ in the long-only case.

Policy-based methods exhibit robust performance even when dealing with noisy datasets, common in stock-related data \cite{MOODY_2001, DU_2009}. Furthermore, Policy-based methods typically converge more swiftly than Value-based methods \cite{GRONDMAN_2012}, although the high variance of the gradient leads to a slower learning rate \cite{KONDA_1999}. These methods inherently perform exploration \cite{PLAAT_2022}, as the stochastic policy yields a distribution over the action space. However, Policy-based methods require a differentiable reward function. Policy-based approaches in finance were pioneered by Moody and Wu \cite{MOODY_1997}, with many variants since then proposed. RRL typically uses a recurrent NN structure, creating dependencies over previous steps and facilitating multi-period optimisation.

\subsection{Actor-Critic Method}

\subsubsection{\textbf{Core Framework for Actor-Critic RL in QF}.}
The Actor-Critic method in RL is a hybrid approach that combines the strengths of Policy-based and Value-based methods to create a robust learning framework. It consists of two main components: the Actor and Critic modules. The Actor module takes the state $S_{t}$ as input and determines the action $A_{t}$ at time $t$. The Critic module subsequently receives the state $S_{t}$ and the action $A_{t}$ determined by the Actor module, evaluates the state-action pair, and computes the reward, adhering to the general framework outlined in subsection \ref{subsec:Critic-Only-method}. Generic Actor-Critic frameworks are discussed in \cite{SUTTON_2018, PLAAT_2022}. The core components of an Actor-Critic framework can be described as follows:
\begin{itemize}
    \item \textbf{Actor:} The Actor is responsible for selecting actions based on the current state. It learns a policy, which is a mapping from states to actions. This policy can be:
    \begin{itemize}
        \item \textbf{Deterministic:} The Actor always chooses the same action for a given state.
        \item \textbf{Stochastic:} The Actor chooses actions according to a probability distribution.
    \end{itemize}
    \item \textbf{Critic:} The Critic evaluates the action taken by the Actor by computing a value function. This value function estimates the expected cumulative reward (discounted over time) of being in a given state and taking a particular action.
    \item \textbf{Advantage Function:} The advantage function helps to determine how much better or worse a particular action is compared to the average action taken from that state. It is defined as:
    \begin{equation}
    A(S_t, A_t) = Q(S_t, A_t) - V(S_t), \label{eq:advantage-function}
    \end{equation}
    where \( Q \) is the action-value function and \( V \) is the state-value function.
    \item \textbf{Gradient Ascent:} Both the Actor and the Critic are trained using gradient ascent. The Actor updates its policy parameters to maximise the expected cumulative reward, while the Critic updates its parameters to provide more accurate evaluations of the actions.
    \item \textbf{TD Error:} Temporal Difference (TD) error is used to update both the Actor and the Critic. It is the difference between the expected reward and the actual reward received, given by:
    \begin{equation}
    \delta_t = R_{t+1} + \gamma V(S_{t+1}) - V(S_t). \label{eq:td-error}
    \end{equation}
\end{itemize}
The integration of deterministic and stochastic policies within the Actor-Critic framework has shown potential for enhancing decision processes across various complex environments. The first application of the Actor-Critic framework in our survey is found in \cite{CHAN_2001}, where the Actor-Critic method is employed to solve the market-making problem (more details, in subsection \ref{subsec:Market-making-and}).
\subsubsection{\textbf{General Observations, Comments and Definitions for Actor-Critic Method}.}

Among the areas covered in this survey, the Actor-Critic method is the least represented (see Table ~\ref{tab:RL Method}). Yet, it remains among the most compelling of the four primary approaches, as it combines the advantages of both Policy-based and Value-based RL methods. As indicated in subsection \ref{subsec:Actor-Only-method}, a notable challenge with Policy-based methods is their \textit{high variance}, which may result in slower convergence or the propensity to become stuck in local optima. The Critic component in Actor-Critic methods mitigates this by providing a value function that stabilises policy-gradient updates, reducing variance. Moreover, Value-based RL methods are prone to \textit{high bias} due to approximation errors in the value function \cite{PLAAT_2022}. The Actor component in Actor-Critic methods helps to directly optimise the policy, reducing bias by ensuring continuous adjustment based on accurate evaluations of actions. Several Actor-Critic algorithms featured in this literature are specifically engineered to surmount these obstacles. The algorithmic frameworks predominantly observed within the literature encompassed by this survey include the Deterministic Policy Gradient (DPG) \cite{SILVER_2014}, Deep Deterministic Policy Gradient (DDPG) \cite{LILLICRAP_2015}, Asynchronous Advantage Actor-Critic (A3C) and the non-parallel version (A2C) \cite{MNIH_2016}, Trust Region Policy Optimisation (TRPO) \cite{SCHULMAN_2015}, Proximal Policy Optimisation (PPO) \cite{SCHULMAN_2017}, Soft Actor-Critic (SAC) \cite{HAARNOJA_2018}.

Though comparatively under-researched in QF, the Actor-Critic category shows significant promise in contemporary applications. Various methods within this category, as documented in works by Mnih et al. \cite{MNIH_2016, LILLICRAP_2015} and Lillicrap et al., \cite{LILLICRAP_2015}, have demonstrated state-of-the-art performance, marking this as a potential area for future QF breakthroughs.

\subsection{Model-Based RL\label{subsec:Model-based RL}}
\subsubsection{\textbf{Core Framework for Model-Based RL in QF}.}

In model-free RL methods, the agent updates a policy directly from the environment's feedback on its actions. On the contrary, in model-based methods involve constructing a model of the environment, which is then used to simulate and plan actions. A foundational framework can be described as follows:
\begin{enumerate}
    \item \textbf{Define a Finite Set of States \(S_t\):}
    States represent environment-derived information at each time point \(t\). This information includes financial accounting data, prices, sentiment, and technical indicators.
    
    \item \textbf{Define a Set of Actions \(A_t\):}
    These are the possible actions of the agent at each time \(t\), such as Buy, Sell, and Hold.
    
    \item \textbf{Learn Transition Probabilities:}
    Construct a transition model that predicts the next state \(S_{t+1}\) and reward \(R_{t+1}\) given the current state \(S_t\) and action \(A_t\). This model can be a neural network or any other function approximator.
    
    \item \textbf{Formulate a Reward Function \(R_t\):}
    Provides numerical feedback to the agent in response to its preceding action, incorporating factors such as profit, risk, and transaction costs.
    
    \item \textbf{Planning and Policy Optimisation:}
    The learnt model can be used to simulate future states and rewards, allowing the agent to plan and optimise actions. Techniques such as Monte Carlo Tree Search (MCTS) or Dynamic Programming can be used for this purpose \cite{PLAAT_2022, SUTTON_2018}.
\end{enumerate}

This framework highlights the core components required for model-based RL in finance, facilitating the creation of a system where an agent learns to make optimal trading decisions through simulation and planning. Model-based RL is the least represented method in the current literature \cite{GUEANT_2019, YU_2019, WEI_2019}.

\subsubsection{\textbf{General Observations, Comments and Definitions for Model-Based RL}.}

Model-based methods in RL offer several advantages and present unique challenges, particularly when applied to financial trading systems: 

\begin{enumerate}
    \item \textbf{Learning Speed:}
    Model-based RL methods typically learn faster than model-free approaches by using the learnt model for planning and action optimisation, crucial for timely financial market decisions.
   
    \item \textbf{Computational Complexity:}
    Simulating and planning with complex financial models is computationally intensive, requiring efficient algorithms and high-performance computing, especially for HFT applications.
    
    \item \textbf{Risk Management:}
    Robust risk management is vital. Model-based RL can simulate extreme market scenarios to assess potential risks, helping to develop strategies that maximise returns and manage risks effectively.
\end{enumerate}

\section{Environment\label{sec:Enviroment}}


In the RL framework, the environment characterises the current state of the system. The agent, the learner, and decision maker interact with this environment, selecting actions based on state information. This setup requires that the agent and the environment are mutually exclusive \cite{SUTTON_1988}, providing distinct boundaries for rewards, actions, and states\footnote{In financial markets, this separation becomes somewhat blurred as an agent's actions—buying and selling assets—can influence the environment. However, most of the current literature assumes that the agent has minimal market impact, a realistic assumption for small investors.}. In financial contexts, the agent, an asset owner, uses the state of financial markets (the environment) and external factors such as stock indices, interest rates, commodity prices, and macroeconomic, political, and natural risks to inform actions (see, Figure \ref{fig:RL_FLowGraph}).

Assuming that the agent can access all relevant information, the RL problem can be addressed under the MDP framework \cite{BARTO_1985}. However, this assumption implies that the environment is fully observable and that future states depend only on the current state and action, a property known as the Markov property. In financial markets, this assumption may be overly simplistic. Empirical evidence suggests that financial markets exhibit longer memory, influenced by factors such as investor behaviour, economic cycles, and external events. These factors introduce dependencies that extend beyond immediate state transitions \cite{CONT_2001, LO_1991}.

Given the complexity and partial observability of financial markets, a Partially Observable Markov Decision Process (POMDP) framework is more appropriate \cite{ASTROM_1965}. In Partially Observable (PO) environments, the transition probabilities between states in financial markets are typically not explicitly modelled. Instead, historical data and statistical methods, such as LSTM or RNN \cite{HEESS_2015}, are used to approximate these transitions. This approach acknowledges the inherent randomness and partial observability of financial markets, making exact environment representation virtually impossible.

Unlike full observability in video games \cite{MNIH_2013, MNIH_2015}, financial markets epitomise a PO environment, lacking deterministic representation; a single day's pricing cannot encapsulate the state of the environment. The use of LSTM and RNN models is relevant here, as they can capture long-term dependencies and patterns in sequential data. These models help address the limitations of the Markov assumption by incorporating memory effects and providing a dynamic representation of the financial markets.

In early adoptions of RL in QF, the environment was assumed to be fully observable \cite{NEUNEIER_1996}, mainly because the inherent randomness of financial markets makes an exact environment representation impossible. Adding more features does not necessarily improve performance \cite{MARTINEZ_2009}, necessitating strategic feature selection to manage randomness, avoid the curse of dimensionality \cite{BELLMAN_1957}, and address interpretability issues. We note that empirical studies have shown that models that incorporate memory effects, such as Long Short-Term Memory (LSTM) \cite{HOCHREITER_1997} and Recurrent Neural Networks (RNN) \cite{RUMELHART_1985, DENG_2016, GERS_2000}, significantly outperform those based on the Markov assumption, particularly in predicting long-term trends and capturing market anomalies \cite{FISCHER_2018_2}.

\subsection{Features\label{subsec:Features}}


Financial market modelling poses significant challenges due to inherent randomness. Thus, the choice and extraction of input data and features are crucial for the effectiveness of RL systems. The literature showcases a variety of data sources and feature selection methods that underline their specific relevance within this domain. Table~\ref{tab:Data_and_Features} (in Appendix) details the features and data from the top fifteen most-cited papers in our sample as of October 2022, listed chronologically.

State representation commonly incorporates discrete state, technical analysis, pricing data, macroeconomic indicators, sentiment data, current position, and Limit Order Book (LOB) data. Few publications experiment with or compare different state configurations. For example, Nevmyvaka et al. \cite{NEVMYVAKA_2006} blended private and market variables, while Briola et al. \cite{BRIOLA_2021} tested three different state scenarios for the RL agent, each with more features.

\subsubsection{\textbf{Price History.}\label{subsec:Price-history}}

Price history and Open-High-Low-Close (OHLC) or OHLCV when incorporating volume \cite{HUANG_2018, THEATE_2021} values are prevalent features in the literature for assets such as stocks or bonds, along with their derivatives, like $20$-day lagged returns \cite{CHEN_2019} or co-integration \cite{ENGLE_1987} based features \cite{LEI_2020}. However, reliance on price-based features should not hinder the consideration of other features. For example, Sherstov and Stone \cite{SHERSTOV_2004} found that using a single price feature underperformed compared to benchmark strategies, hinting at the need for a richer state representation to capture the complexities of the financial market. Moreover, Benhamou et al. \cite{BENHAMOU_2020_A} noted the high correlation between OHLC features, which could introduce input noise. Thus, feature selections on raw features might be a more effective strategy in certain contexts, as discussed in subsection \ref{subsec:Features-selection}. 

Volatility, a critical feature derived from historical prices, is surprisingly overlooked in the literature, despite its crucial role in PM \cite{MARKOWITZ_1952} and in identifying shifts in financial markets (regime changes) \cite{LEBARON_1992}. Its absence in RL-based PM applications \cite{YU_2019, YE_2020, LIU_2020} is noteworthy given its importance \cite{BENHAMOU_2020_A}. Volatility has only recently been used in PM applications \cite{BENHAMOU_2020_A,BENHAMOU_2020_B,BENHAMOU_2021_A,BENHAMOU_2021_B}.
In other RL-based finance applications, different volatility measures have been successfully used as a tool to discover regime changes \cite{MARINGER_2010, MARINGER_2012, MARINGER_2014}. In effect, the authors extended the original work of Moody and Wu \cite{MOODY_1997} by adding a regime-switching extension to the RRL framework. In Bekiros \cite{BEKIROS_2010}, a comparison is made between changes in 20-day volatility and the previous day, while Tan et al. \cite{TAN_2011} used the standard deviation of a stock price and its correlation with the Dow Jones index to identify stock-specific cycles. Zhang and Maringer \cite{ZHANG_2016} included Garch \cite{BOLLERSLEV_1998} volatility in state representation, marking its first use in an Policy-based approach. Finally, certain authors incorporate the full covariance of closing prices in state modelling \cite{KATONGO_2021}.

\subsubsection{\textbf{Technical Analysis.}}

Technical analysis employs indicators and rules to predict price directions based on past price and volume data \cite{MURPHY_1999}. The validity of these methods is frequently contested by the Efficient Market Hypothesis (EMH), which posits that stock prices are inherently unpredictable \cite{FAMA_1970}. Despite mixed research outcomes and some methodological criticisms \cite{PARK_2007}, numerous scholars challenge the EMH by suggesting that past prices can forecast future trends, thereby refuting the random walk hypothesis \cite{LO_1988}. Consequently, technical indicators such as Moving Average (MA), Exponential Moving Average (EMA), Moving Average Convergence/Divergence (MACD), Japanese candlestick, and Relative Strength Index (RSI) are widely used to represent environments within Reinforcement Learning (RL), as evidenced in various studies (e.g., \cite{DEMPSTER_2001, DEMPSTER_2002, DEMPSTER_2006, GABRIELSSON_2015, PARK_2020, WU_2020, ZHANG_2020_A, WANG_2021_B}). However, their efficacy in RL applications remains a subject of debate \cite{DEMPSTER_2006}.




\subsubsection{\textbf{Fundamental Data and Factor Investing.}\label{subsec:Fundamental-data-and}}

Accounting data underpins traditional factor investing, with numerous strategies \cite{HARVEY_2019} developed over decades to explain expected asset returns \cite{COCHRANE_2005}. Key factors include price momentum\footnote{Trading strategies where one buys winner stocks that performed well and sells those that did poorly in the past to generate profits over a 3-12 months holding horizon.} \cite{JEGADEESH_1993}, value\footnote{Buying stocks with relatively low prices based on fundamental ratios and following the opposite when the high share prices are detected.} \cite{FAMA_1992}, size\footnote{Small companies perform better than big companies based on some measure, such as market capitalisation.} \cite{BANZ_1981}, quality\footnote{Buying good quality companies and selling poor quality companies based on information reported in the financial statements.} \cite{GRAHAM_1973}, and low beta\footnote{Buying stocks with low risk and selling stocks with high risk.} \cite{FRAZZINI_2014}. Despite their proven efficacy, these factors are infrequently used in RL due to the quarterly updates of accounting data, limiting their effectiveness as daily signals. Nevertheless, pioneering work by Zhang and Maringer \cite{ZHANG_2013,ZHANG_2016} and recent efforts by Wang et al. \cite{WANG_2019} and Coqueret and André \cite{CORQUERET_2022} have begun integrating these factors into RL environments, potentially enhancing the models' predictive capabilities.

\subsubsection{\textbf{Order Book.}}
Among the literature surveyed, we find a substantial focus on the areas of trade execution and market making. These specific applications tend to employ a unique set of features to model the environment, distinguishing them from other domains. While we explore these applications more comprehensively in subsections \ref{subsec:Order-execution} and \ref{subsec:Market-making-and}, it is beneficial to highlight some of the most commonly observed features in the associated literature. For example, Chan and Shelton \cite{CHAN_2001} used order imbalance-based features, depth of the market, and time to fill a limit order, among others. Other authors used the Bid-Ask spread and remaining inventory, among others, as part of their environment \cite{NING_2020}. Numerous other features are also deployed in this context, but a detailed discussion is omitted here for brevity.
\subsubsection{\textbf{Sentiment Data.}}
The impact of sentiment data on stock prices is well documented in the financial literature \cite{ANTWEILER_2004, TETLOCK_2007}. Its application in RL strategies has also been demonstrated \cite{FEUERRIEGEL_2016, KAUR_2017, YANG_2018, NAN_2020, YE_2020, SAWHNEY_2021, KORATAMADDI_2021, HUANG_2022}. Diverse sources like Reuters News Corpus, Twitter, and Thomson Reuters News analytics have proven effective in creating sentiment signals that reduce uncertainty in financial models. However, the application of sentiment data in RL is predominantly focused on well-known companies, thus limiting its broader applicability to lesser-known stocks. Despite its demonstrated value, sentiment data is underutilised in state representations across the existing RL-based financial models.
\subsubsection{\textbf{Macroeconomic.}}
Macroeconomic data, including widely recognised indicators such as interest rates, inflation, and GDP, plays a crucial role by reflecting the broader economic context in which companies operate, particularly for cyclical firms closely aligned with economic fluctuations. However, the integration of such data into state representations in RL models is rare. Neuneier's work \cite{NEUNEIER_1996} was pioneering in incorporating interest rates \cite{FISCHER_2018}, and Benhamou's subsequent studies \cite{BENHAMOU_2020_A, BENHAMOU_2020_B, BENHAMOU_2021_A} expanded this approach by including the slope of the US Treasury yield curve as a risk indicator.
\subsubsection{\textbf{Others.}}
In earlier subsections, we covered several financial features which can be or are used in the current RL literature. The selection and use of these features often depend on the specific application, resulting in bespoke adaptations. This section consolidates a few of those unique variations. In trade execution, typical state representations include elapsed time, remaining shares to order \cite{NEVMYVAKA_2006}, and current time \cite{NING_2020}. Deng et al. \cite{DENG_2015} used over 80 raw features, such as the volume-weighted average price (VWAP) and credit spreads of global corporate bonds \cite{BENHAMOU_2020_A, BENHAMOU_2020_B, BENHAMOU_2021_A, BENHAMOU_2021_B}. Other studies have incorporated current and cash positions into their models \cite{SHERSTOV_2004, BERTOLUZZO_2012, JIN_2016, JIANG_2017, JIANG_2017_1, KAUR_2017}. Additionally, emerging data sources such as Environmental, Social, and Governance (ESG) factors \cite{GOMPERS_2003, OIKONOMOU_2012} and supply chain data \cite{COHEN_2008, MENZLY_2010} could be potential candidates for integration into RL environments to enrich the understanding of market dynamics and firm interconnections.
\subsection{Feature Selection and Feature Extraction\label{subsec:Features-selection}}
The complexity of financial markets necessitates numerous features to define an agent’s environment, which can increase dimensionality and computational demands. Early Value-based applications addressed \cite{DEMPSTER_2002} this by discretising states, such as binarising technical indicators into bit strings \cite{HOLLAND_1992}. Significant advancements in feature selection and representation  occurred after 2016, driven by improvements in DRL and the adoption of neural network (NN) architectures such as Convolutional Neural Networks (CNN) \cite{MNIH_2013, MNIH_2015, LECUN_1995, KRIZHEVSKY_2012}. We discuss here the mechanisms for feature selection and extraction in financial RL.

\subsubsection{\textbf{Deep Learning Based Features.}}

RNNs are a vital tool in POMDPs due to their ability to record and use past states to optimise future actions \cite{RUMELHART_1985, DENG_2016, GERS_2000}. However, vanishing gradients in deep structures are an issue \cite{BENGIO_1994}. Consequently, many researchers have shifted focus to LSTMs in the reviewed literature. The use of LSTM in decision-making is advantageous due to its ability to remember features over longer steps, an essential quality for financial market trading. Additionally, its memory cell retains trading actions, thereby capturing the impact of transaction costs more effectively. For example, Bisht and Kumar \cite{BISHT_2020} developed two LSTM-based systems, one functioning as an autoencoder for feature extraction and the other using the decoded features within an LSTM module for decision-making. Other notable applications of LSTM include those found in \cite{LU_2017, CHEN_2019, JIA_2019, PONOMAREV_2019, TSANTEKIDIS_2020, LIN_2020, SAWHNEY_2021, KUMAR_2022}.

Transitioning from RNNs and LSTMs, CNNs introduce a different paradigm in NN architecture. CNNs are a crucial aspect of feature extraction in PM due to their ability to independently analyse asset price movements and manage complex data structures. Early implementations \cite{JIANG_2017, JIANG_2017_1} introduced the Ensemble of Identical Independent Evaluators, which independently extracted the price movement of each asset. Challenges such as vanishing gradients and the inability to capture non-linear relationships prompted a transition towards Deep Residual Networks \cite{HE_2016} for enhanced robustness \cite{LIANG_2018}. Further advancements addressed the lack of temporal dynamics in financial models by employing the Inception Network for multi-scale analysis \cite{SZEDEGY_2015}, integrating multiple data aspects into a cohesive framework \cite{SHI_2019}. Additionally, innovations like the integration of autoencoders and dual CNNs have expanded boundaries by enhancing dimensionality reduction and combining time series with cross-sectional data analysis in an Actor-Critic framework, marking significant progress in the domain \cite{SOLEYMANI_2021, ABOUSSALAH_2022}. These applications underscore the evolving capabilities of CNNs to provide comprehensive insights into financial asset interactions. Further contributions to the CNN literature are evidenced by additional studies \cite{TAGHIAN_2022, BENHAMOU_2020_A, BENHAMOU_2020_B, BENHAMOU_2021_A, BENHAMOU_2021_B}.

The integration of attention mechanisms marks yet another advance in NN technology, significantly enhancing NN in QF by improving temporal relationship analysis and long-term dependency modelling \cite{SUTSKEVER_2014, VASWANI_2017, PEI_2017}. These mechanisms have been successfully applied in various settings to boost the performance of financial models, as evidenced by their use in diverse applications across recent studies \cite{LEI_2020, WENG_2020, ZHANG_2020_B, WANG_2019, WANG_2021_A, XU_2021}

Lastly, innovative approaches like Generative Adversarial Networks (GANs) \cite{GOODFELLOW_2014, ESTEBAN_2017}, Deep Belief Networks (DBNs) \cite{HINTON_2006}, and Gate Recurrent Units (GRU) \cite{CHO_2014} have been used in the reviewed literature for data augmentation \cite{YU_2019} and for feature extraction are addressing the scarcity of extensive financial datasets \cite{ABDELKAWY_2021, WU_2020, LIU_2020}.

\subsubsection{\textbf{Other Feature Mechanisms.}}\label{subsec:Others-feature-mechanisms}

Yuan et al. \cite{YUAN_2020} use minute-candle data to train their RL model for daily stock trading, proposing a skewness and kurtosis-based selection process for trading stocks. Several works use OCHLV data in state representation without theoretical or practical explanations \cite{JIANG_2017,JIANG_2017_1}. However, Weng et al. \cite{WENG_2020} provide some clarity by revealing through XGBoost \cite{CHEN_2016} that the closing price is the most important feature of all. Clustering methods for simplifying state representation are used in studies such as Fengqian and Chao \cite{FENGQIAN_2020}, who use Japanese candlesticks for denoising. Other examples include \cite{CHAKOLE_2021, KIM_2022}.

Various transformations and techniques are proposed to reduce the inherent randomness and uncertainty of financial data. Deng et al. \cite{DENG_2016} suggest a fuzzy transformation \cite{LIN_1991}, while others apply sparse coding~\cite{DENG_2015}, and Carta et al. \cite{CARTA_2021} use the Gramian Angular Field (GAF) \cite{WANG_2015} to transform time series into images. Lu et al. \cite{LU_2022} recommend the Continuous Wavelet Transform (CWT) \cite{RIOUL_1992} to derive frequency-domain representations, which are seen as more suitable for time-varying signals, which they integrate with a CNN and LSTM to extract frequency- and time-domain features. Dimensionality reduction is also used to retain salient features \cite{LI_2021, KATONGO_2021, QUI_2021}. Carapuço et al. \cite{CARAPUCO_2018} adhere to the data preprocessing guidelines established by LeCun et al. \cite{LECUN_2002}.

Genetic Algorithms (GA) have been widely used in RL for different tasks. These include selecting the best features, optimising how models are evaluated, using extra data like order flow and order books, choosing the right technical indicators, and improving trading strategies. Several studies have explored these uses (e.g., \cite{DEMPSTER_2002, BATES_2003, HRYSHKO_2004, CHEN_2007, MABU_2007, GU_2011, ZHANG_2013, MARINGER_2012, ZHANG_2016, ZHU_2018}).
 
\section{Action Modelling and Reward Functions in Finance\label{sec:Action-Modelling-and}}



In this section, we emphasise two pivotal facets of the RL framework: the action space and reward functions tailored to finance. Actions, denoted \(A_{t}\), represent the agent's interactions with the environment at time \(t\). In QF-related RL studies, agents typically do not affect the environment, with actions ranging from discrete (e.g., buy or sell decisions) to continuous (e.g., adjusting portfolio weights). Table \ref{tab:Actions-and-Rewards} (in Appendix) presents the action and reward functions from the top fifteen cited papers in our survey.

The agent's objective, encapsulated by the reward signal $R_{t} \in \mathcal{R \subset \mathbb{R}}$, optimises the expected cumulative reward over a sequence of steps, underscoring the preference for long-term gains. The nature of these rewards is application-dependent, commonly aligned with financial objectives such as maximising the SR or cumulative wealth. This objective corresponds to the return $G_{t}$, defined as the discounted sum of sequential rewards $R_{t+1},R_{t+2},R_{t+3}...$, formalised as:
\begin{equation}
G_{t} = \sum_{k=0}^{\infty}\gamma^{k}R_{t+k+1},\label{eq:-15}
\end{equation}
where \(\gamma \in [0,1)\) represents the discount rate determining the agent's foresight. With \(\gamma = 0\), the agent prioritises immediate rewards, whereas \(\gamma\) close to 1 emphasises future rewards significantly. Notably, for infinite reward sequences, \(\gamma \in (0,1)\) ensures equation (\ref{eq:-15}) remains finite and mathematically well-defined, which is crucial for continuing tasks. 

Drawing parallels with finance, this approach conceptually resembles the discounting of the price of bonds, where \(\gamma\) is the period discount factor and \(R_{i}\) represents the coupon plus the face value at time \(i\) \cite{FABOZZI_2012}. In finite-time horizons, such as modelling the expected return until a terminal event such as bankruptcy, using a discount factor \(\gamma < 1\) is consistent with financial models such as the Gordon Growth Model \cite{GORDON_1959} and bond pricing. These models inherently apply a discount factor of less than 1 to reflect the time value of money, ensuring that the present value of future cash flows is properly calculated. This method emphasises the financial principle that future rewards are typically less valuable than immediate rewards, even within a finite horizon. However, setting $\gamma = 1$ can be appropriate and meaningful in specific bounded scenarios where future rewards are equally critical.

\subsection{Action\label{subsec:Action}}

Action spaces in RL are typically either continuous or discrete, usually reflected in the agent's framework. While Value-based methods exclusively accommodate discrete action spaces, alternative methodologies exhibit greater flexibility, rendering them more realistic for financial trading applications. However, the utility of Value-based methods should not be discounted, as the following discussion in subsection \ref{subsec:PortManagement} will reveal.

Most Value-based RL studies in QF allow for two to three actions (Buy/Hold/Sell) for a single stock. Exceptions exist, such as Sherstov and Stone's \cite{SHERSTOV_2004} model encompassing 1801 actions and Dempster and Romahi's \cite{DEMPSTER_2002} bitstring-coded state representation. Similarly, models with multiple tradable assets, such as Kaur's \cite{KAUR_2017} model with three assets, demand a wider range of actions, thereby escalating computational time due to an expanded exploration scope. Jangmin et al. \cite{JANGMIN_2006} and Park et al. \cite{PARK_2020} suggest alternatives to these limitations by using multiple local traders and creating an ETF portfolio, respectively. However, even these sophisticated models employ discrete action spaces. Recently, research has focused on addressing this limitation in DRL, such as Kalashnikov et al.'s \cite{KALASHNIKOV_2018} QT-Opt method, employing the Cross-Entropy Method (CEM) for Q-function optimisation.

Despite such advances, Value-based methods still lag behind Policy-based and Actor-Critic methods in terms of efficiency, especially when dealing with continuous action spaces. This ability is crucial in financial markets, where action spaces are naturally continuous. Consequently, using RL algorithms such as SARSA and Q-learning for continuous actions can lead to suboptimal solutions or even failure to converge, as observed in some simple cases \cite{BAIRD_1998}. 

In the RRL framework, the action space can be represented as an NN where the state inputs translate into actions \cite{GOLD_2003}. In single-asset trading systems, typically a single output neuron (activated by a $\tanh$ function) is used, the output of which models long and short positions. Such output can be discretised into buy/hold/sell signals with defined thresholds. For multi-asset scenarios, such as in  Jiang et al. \cite{JIANG_2017,JIANG_2017_1}, the softmax activation function is used to ensure the sum of portfolio weights equals one. Variations of this approach are seen in the literature, where there is a distinction between long and short portfolios \cite{WANG_2019, WANG_2021_A}. However, Coqueret and André \cite{CORQUERET_2022} suggest being cautious against the indiscriminate use of the softmax function due to its potential violation of the budget constraint. Instead, they advocate for Dirichlet distributions, apt for long-only portfolios due to inherent budget adherence. In comparison, Actor-Critic methods align with Policy-based in their action set, albeit with some unique implementations. For example, Li et al. \cite{LI_2007} harnessed agent action to improve SVM price prediction, while Bekiros \cite{BEKIROS_2010} proposed a trading system that intertwines agent action with a fuzzy system. Qiu et al. \cite{QUI_2021} further innovated by introducing Quantum Price Levels (QPL) to model the action space.

\subsection{Rewards \label{subsec:Rewards}}

The selection of the reward function in RL is crucial, as it indicates the objective to be optimised \cite{SUTTON_2018}. The flexibility inherent in RL enables the creation of a wide range of reward functions tailored to specific tasks or problems. Tan et al. \cite{TAN_2011} demonstrate this by introducing a range of task-specific reward functions.

Several studies contrast different reward functions and frameworks, such as Du et al. \cite{DU_2009}, who analyse various reward functions, including cumulative wealth, utility, and SR in the context of Q-learning and RRL, highlighting their interconnection. Although utility-based rewards offer a simple way to infuse risk sensitivity into the objective function, Mihatsch and Neuneier \cite{MIHATSCH_2002} warn of inherent drawbacks. They suggest a risk-sensitive RL approach using transformed temporal differences during learning and propose a novel algorithmic variation of temporal difference and Q-learning. Lucarelli et al. \cite{LUCARELLI_2019} indicate that Sharpe Ratio (SR) reward functions outperform profit-based reward functions in DQN contexts. Finally, immediate rewards tend to be the primary focus in the literature, because not selecting the immediate reward will incur additional transaction costs in quantitative trading.

\subsubsection{\textbf{Profit and Performance-Based Rewards.}\label{subsec:Profit-and-performance}}

Most academic research gravitates towards profit-based rewards \cite{CHAN_2001, EILERS_2014, KAUR_2017, SI_2017, SORMAYURA_2019, SAWHNEY_2021}. Risk Performance Measures (RPM), such as SR, comprise the second most common focus \cite{BERTOLUZZO_2012, CORAZZA_2015, WANG_2019, THEATE_2021}). An intriguing variation emerges in several studies \cite{MARINGER_2014, ZHANG_2014, GABRIELSSON_2015, LIU_2020, LI_2021}, notably the use of Differential SR (DSR) as introduced and developed further by Moody and colleagues \cite{MOODY_1997, MOODY_1998, MOODY_1998_2, MOODY_2001}. DSR, an online variant of standard SR, offers several benefits, including enabling efficient online optimisation and weighing more recent returns \cite{MOODY_1997}. The generic derivation of the DSR suggests the feasibility of online versions for other performance measures \cite{MOODY_1997, BERTOLUZZO_2007, ALMAHDI_2017, ALMAHDI_2019}. Alternative measures have been used in various contexts, for instance, Carapuço et al. \cite{CARAPUCO_2018} and Wu et al. \cite{WU_2020} use the Sortino ratio \cite{SORTINO_1994}. The diversity of these measures highlights the potential to explore broader reward functions, such as incorporating higher moments into the classic SR \cite{PEZIER_2006}.


Several studies have proposed unique risk-averse reward functions. For example, Jin and El-Saawy \cite{JIN_2016} define a reward function that incorporates both return and standard deviation, a structure echoed in Li and Chan \cite{LI_2006}, who instead use GARCH \cite{BOLLERSLEV_1998} volatility. Si et al. \cite{SI_2017} and Bisht and Kumar \cite{BISHT_2020} incorporate downside volatility, reflecting a more practitioner-orientated focus. Such risk-aware reward functions better capture the priorities of portfolio managers who balance risk and return.

\subsubsection{\textbf{Sparse and Conditional Rewards.}}

Sparse rewards present a realistic model of many environments, typically characterising a reward function with predominantly zero values and few positive state-action pairs. Although sparse rewards offer a detailed description of various environments, they can pose challenges in identifying optimal policies due to the infrequent positive feedback \cite{PLAAT_2022}. Another form of reward function similar to sparse rewards is a conditional reward, in which the reward depends on specific conditions in the state or action.  Usually, we observe sparse rewards or conditional-based rewards in cases where the action space is discrete, such as BUY, SELL, and HOLD in the Value-based framework.

Neuneier \cite{NEUNEIER_1996} first used conditional rewards within a foreign exchange context. The state vector $s_{t}$ comprises the current exchange rate $x_{t}$, portfolio wealth $c_{t}$, and a binary variable $b_{t}$ that indicates the investment in DM (Deutsch Mark) or USD at time $t$. The reward function, then defined at $t+1$ given the action, integrates the transaction costs into the model, thus reducing the reward for a DM-USD transition. Numerous studies have expanded on this framework. Jia et al. \cite{JIA_2019} designed a sparse reward incorporating market price, volume, transaction costs, and the current amount. Kim and Kim \cite{KIM_2019} used conditional rewards for risk management in statistical arbitrage. Lucarelli et al. \cite{LUCARELLI_2019} introduced a hybrid reward function based on discretised and thresholded SR values, leading to specific Sharpe Ratio (SR) ranges that increase or decrease the reward. Lin and Beling \cite{LIN_2020} employed a sparse reward for optimal trade execution based on performance relative to Time-Weighted Average Price (TWAP). Similarly, Carta et al. \cite{CARTA_2021} incorporated conditional rewards based on market prices, introducing a zero-reward state for idle agents. Taghian et al. \cite{TAGHIAN_2022} applied a similar framework in their trading systems.

In this context, sparse and conditional rewards generally serve to control extreme reward values \cite{KIM_2019} or as a mechanism to alter trading signals \cite{NEUNEIER_1996}, thus shaping the learning trajectory towards optimal policy discovery.

\subsubsection{\textbf{Utility-Based Rewards.}\label{subsec:Utility-based-rewards}}
Utility-based rewards have an established presence in the RL finance literature, initially presented by Dempster and Leemans \cite{DEMPSTER_2006}, where they optimised an RL-based trading system according to a utility function. This utility function resembles the principles of classical portfolio construction, with the aim of generating higher returns for a given level of risk. The maximisation of the utility depends on several parameters, five in total. Subsequent studies expanded on these principles. Hens and Wöhrmann \cite{HENS_2007} applied the power utility 
function for dynamic allocation between bonds and equity indices. In a shift towards high-frequency market making, Lim and Gorse \cite{LIM_2018} used Constant Absolute Risk Aversion (CARA) to gauge an agent's attitude towards gains or losses on partial inventory sales.

García-Galicia et al. \cite{GARCIA_2019} proposed an Actor-Critic framework implementing a utility-based reward function for continuous-time PM. Bao and Liu \cite{BAO_2019}, as well as Bao \cite{BAO_2019_B}, employed the difference between two successive utility functions as a reward. Risk-averse agents and ES/CVAR, the average loss exceeding the worst $q\%$ of cases \cite{ROCKFELLAR_2000}, are also adopted as a risk-adjusted performance measure~\cite{BUEHLER_2019, ALAMEER_2021}. Finally, in market-making and trade execution literature, utility-based rewards are common \cite{NING_2020, KOLM_2020, SPOONER_2020}.

\subsubsection{\textbf{Composite Reward Signals.}}

Several studies employed multiple reward functions to optimise agent behaviour. An intuitive and straightforward approach is the linear combination of disparate reward functions. This practice can address the issue of reward noise, which is particularly prevalent in profit-based signals. When rewards contain high noise levels, they can induce substantial variation in returns and thus obstruct the training process. Furthermore, it becomes necessary to normalise the rewards when they span different scales. Employing composite rewards offers a natural avenue for integrating a multi-objective structure within our RL agent, although caution must be exercised with potentially conflicting rewards.

This type of reward system was first developed within Chan and Shelton's market-making framework \cite{CHAN_2001}. They identified three distinct reward signals—changes in profit and inventory levels—and integrated them linearly. The training process determined the weights assigned to each signal. More recently, Zarkias et al. \cite{ZARKIAS_2019} and Tsantekidis et al. \cite{TSANTEKIDIS_2020} proposed a combined reward signal based on three individual signals: profit and loss, price trailing (the extent to which the asset price is tracked) and fees. They found that their approach functioned as a potent regulariser, enhancing performance in comparison to the unadorned P\&L reward. In Zhang et al. \cite{ZHANG_2020_B}, the final reward was a linear combination of the risk-sensitive and the cost-sensitive reward mirroring Jin and El-Saawy \cite{JIN_2016}). Lastly, Koratamaddi et al. \cite{KORATAMADDI_2021} produced a final reward encompassing a change in portfolio value from the previous day and market sentiment, marking the first instance of market sentiment being integrated into the reward function.

\subsubsection{\textbf{Others.}}

Yang et al. \cite{YANG_2020} present a reward function incorporating the turbulence index \cite{KRITZMAN_2010}. In their methodology, if the turbulence index exceeds a particular threshold, agents act to forestall potential adverse shifts in portfolio value by liquidating all held stocks. Meanwhile, Wang et al. \cite{WANG_C_2021} have adopted a reward-shaping approach. Here, a penalty is appended to a profit-based reward if the agent's actions deviate from those of an expert, baseline, or any other valid comparator, where the expert policy serves as a threshold. The idea to integrate a dynamic strategy, such as an oracle, offers an intriguing prospect for further investigation.

\section{Enhancing RL in Quantitative Finance with Advanced ML Techniques \label{sec:ML-Themes-and}}


In this section, we explore salient ML themes evident in the current RL literature, highlighting their relevance and potential to significantly augment RL applications in QF. RL models can achieve enhanced performance, robustness, and adaptability in financial markets by integrating ensemble methods, transfer learning, imitation learning, policy distillation, and Multi-Agent Systems (MAS). These methods tackle challenges such as data scarcity and complex decision-making environments.

\subsection{Ensemble approaches}

Ensemble trading systems are conspicuously underrepresented in the reviewed literature. Notable studies include \cite{YANG_2020, LEEM_2020, CARTA_2021, KUMAR_2022}. This section highlights promising ideas for integrating ensemble methods to enhance RL-based trading systems in QF.

Yang et al. \cite{YANG_2020} constructed an ensemble consisting of the PPO, A2C, and DDPG algorithms. Specifically, they trained these three agents with an identical growing window of $n$ months. Subsequently, the agent exhibiting the best performance (evaluated using the SR) during the validation test was employed to forecast the trade for the next quarter. This strategy was inspired by their observation that different agents excel under various market conditions. Notwithstanding its simplicity, this approach could be improved by implementing a forward-looking measure for agent selection or adopting multiple performance criteria. Furthermore, understanding why specific agents outperform others under certain market conditions warrants further exploration. In their study, Leem and Kim \cite{LEEM_2020} introduced a DQN ensemble trading system built on three trading experts, each associated with an individual action. A unique reward value was assigned to each action under specified conditions to drive specific investment behaviour. Carta et al. \cite{CARTA_2021} developed a reward-based classifier to process outputs or features extracted from a preceding layer using CNN features. This process is based on stacking \cite{WOLPERT_1992}, which is an ensemble method. Kumar et al. \cite{KUMAR_2022} adopted a parallel framework, with their ensemble method using various feature extraction mechanisms such as CNNs and LSTMs. The resulting agents were then combined via a trio of methods: highest probability, average, and majority voting.

Ensemble methods enhance accuracy \cite{HANSEN_1990}, generalisability \cite{POLIKAR_2006}, and robustness to outliers \cite{OPITZ_1999}, making them ideal for RL-based trading systems in QF. However, they also decrease transparency, increase complexity, and increase computational costs \cite{POLIKAR_2006, ZHOU_2012}. This trade-off between interpretability and benefits requires further investigation.

The construction of reward-based ensembles represents an intriguing frontier for future research. By facilitating the concurrent optimisation of multiple reward functions, this strategy can adapt to the intricate landscapes of RL environments found in QF applications, where performance is evaluated using distinct reward functions. This approach bears conceptual similarities with the multi-agent RL approach, which also centres on the optimisation of multiple agents, each guided by a unique reward function.
\subsection{Transfer learning\label{subsec:Transfer-learning}}
In the past decade, ML has made significant strides in classification and regression tasks. However, these tasks often assume that training and testing datasets share the same distribution. When distributions shift, models typically need rebuilding with new data, which is inefficient. Transfer learning addresses this by using pre-trained models as starting points. In finance, transfer learning is crucial for improving RL agent performance by leveraging pre-trained models to address data scarcity as comprehensively discussed in seminal surveys by Pan et al. \cite{PAN_2010} and Zhuang et al. \cite{ZHUANG_2020}. Transfer learning has profound implications in the realm of finance due to its potential to facilitate the generalisation of experience—a subject we discuss in this section. However, our review of the RL-based literature indicates that its adoption has been relatively limited \cite{JEONG_2019, WEI_2019, HUANG_2022, BORRAGEIRO_2021, BORRAGEIRO_2022}.

Jeong et al. \cite{JEONG_2019} applied DQN-based transfer learning to stock trading, focusing on equity stock indices. They used two techniques to filter the stocks: the correlation between two stocks and an NN to determine the relevance of the data pattern. They categorised the stock data into three groups based on their association with the underlying stock index (measured by lower mean square error (MSE) for NN). Half of the stocks with high correlation and low MSE were used to pre-train a model with shared weights, which was subsequently used to train the main model on the index. They argued that transfer learning compensates for financial data scarcity by using index constituents as proxies, although the opposite can also be anticipated since indices generally have more historical data. Motivated by the sample efficiency and transferability of model-based RL \cite{PLAAT_2022}, Wei et al. \cite{WEI_2019} introduced a model-based RL agent for electronic trading, employing an autoencoder to extract latent LOB features and combining an RNN with a Mixture Density Network (MDN) \cite{BISHOP_1994}. This represents the first application of model-based RL in transfer learning. However, the challenges in modelling financial market environments, particularly state transition modelling, might explain the scarcity of such applications in the literature.

Huang and Tanaka \cite{HUANG_2022} introduced transfer learning to avoid training multiple DQN agents from scratch. They trained a DQN agent on Apple Inc. prices before training others. Although this approach might help understand stock trends, it overlooks that Apple's success may not represent typical stock patterns, particularly regarding default risk, potentially leading to negative transfer \cite{PERKINS_1992}. Borrageiro et al. \cite{BORRAGEIRO_2021} used transfer learning with Radial Basis Functions for feature representation in an RRL framework. Later, they extended this to digital assets, using Echo State Networks \cite{JAEGER_2002} to represent feature space and then transferred this knowledge to RRL agents for Bitcoin trading.


Transfer learning, with its ability to leverage pre-existing knowledge, can significantly boost RL agents' performance in finance. However, some studies could have benefitted from a more strategic application of transfer learning. For instance, rather than using a single stock as in Huang and Tanaka \cite{HUANG_2022}, a sector-based approach might be more effective. Stocks within a sector tend to share similarities, and sector indices often provide more historical data. An optimal strategy might involve training initial models on sector indices and then applying transfer learning to individual stocks within the sector.

\subsection{Policy distillation}

In subsection \ref{subsec:Transfer-learning}, we examined transfer learning in QF. A related concept, Policy Distillation (PD), passes knowledge from a larger model to a smaller one for the same task, hence 'distillation' \cite{HINTON_2015}. Despite its potential to improve the efficiency of the RL agent and mitigate overfitting, PD has been underused in finance, with notable applications by Tsantekidis et al. \cite{TSANTEKIDIS_2021} and Fang et al. \cite{FANG_2021}. PD can help mitigate overfitting and streamline the deployment of RL agents by transferring knowledge from complex models to simpler and more efficient ones.

Tsantekidis et al. \cite{TSANTEKIDIS_2021} proposed a PPO-based agent for foreign exchange trading using PD. They trained "teacher" agents across different currency groups, creating a diversified learning pool. The ``students'', learning a specific data subset, then used PD. To handle financial data noise, the authors employed diversified teacher ensembles. Similarly, Fang et al. \cite{FANG_2021} used PD under PPO for order execution \cite{RUSU_2015}. The aim was to bridge the gap between imperfect market information and optimal order execution policy. Here, the teacher had access to perfect information, aiming to learn the optimal policy while the student mirrored the teacher's behaviour. Despite a few authors applying PD, its potential use cases mirror transfer learning's breadth. A potential comparison or integration with an agent using transfer learning could be beneficial, as PD can fill in missing values.

\subsection{Imitation learning\label{subsec:Imitation learning}}

Imitation Learning (IL), also known as learning from demonstrations, is a machine learning technique that allows an agent to learn a task by imitating expert behaviour. In the QF context, IL has gained attention due to its potential to replicate successful investment strategies. In RL, IL enables agents to mimic expert investors, thereby improving performance and enhancing the accuracy of financial models. The application of IL in QF was explored by Ding et al. \cite{DING_2018}, and later by Yu et al. \cite{YU_2019}, Liu et al. \cite{LIU_2020}, and Kim et al. \cite{KIM_2022}. Finally, the current literature classifies IL methods into two categories: passive and active \cite{ABBEEL_2004, ROSS_2011}.

Ding et al. \cite{DING_2018} developed three investor imitation frameworks—Oracle, Collaborator, and Public Investor—each requiring a distinct reward structure. The Oracle framework leveraged evaluation scores to maximise historical rewards, but it lacked foresight. The Collaborator and Public Investor frameworks instead guided agent learning through a similarity score function and revenue curve approximation, respectively. Yu et al. \cite{YU_2019} deployed an active IL algorithm within a model-based RL framework, employing one-step greedy actions to optimise immediate returns. This methodology involved slight weight perturbations of the actor based on the log loss between the expert's and the actor's actions. Liu et al. \cite{LIU_2020} created a trading-expert algorithm that operated long and short positions at optimal prices. They estimated the gap between agent and expert actions using Ross and Bagnell's method \cite{ROSS_2010}, followed by applying a modified policy gradient to the actor. Meanwhile, Kim et al. \cite{KIM_2022} used behaviour cloning in statistical arbitrage, facilitating learning from a prophetic expert.

IL has significant potential in QF, for example, enabling investors to replicate successful mutual fund performances. However, correct feature selection remains crucial. For instance, imitating a multifactor fund necessitates prioritising features such as value, momentum, low beta, size, and quality. Integrating IL with transfer learning offers exciting prospects for advancing finance models and addresses challenges with limited expert demonstrations. Using transfer learning, the knowledge gained from expert investors can be effectively transferred to different financial markets or scenarios, enabling faster adaptation and improved decision-making. 

\subsection{Multi-agent applications of RL in finance\label{subsec:Multi-agent-applications}}

A novel RL framework for Multi-Agent Systems (MAS) in QF was first proposed by Lee and Jagmin \cite{LEE_2002} and refined in subsequent studies \cite{LEE_2002_2,LEE_2003,LEE_2007}, demonstrating how cooperative agent environments can optimise stock trading solutions. This framework structures the problem so that agents work together towards an optimal solution for stock trading. Lee et al. \cite{LEE_2007} proposed two types of agents: the signal agent and the order agent, each further divided into two agents, forming four agents:

\begin{enumerate}
\item \textbf{Buy signal}: This agent assesses recent price movement to gauge potential price rise. If no buy signal, the episode ends.
\item \textbf{Buy order}: Activated by a buy signal, this agent sets the limit price for buying based on intraday prices.
\item \textbf{Sell signal}: This agent monitors profit and loss to decide when to close a position.
\item \textbf{Sell order}: Activated by a sell signal, sets the limit price for selling based on intraday prices.
\end{enumerate}

This approach recognises that buying and selling stocks involve unique considerations: buyers assess potential price rises or falls, while sellers consider price fluctuations and potential profits or losses -- leading to distinct state representations for each agent. The buy agent's state is informed by long-term price changes, while the sell agent priorities the current profit or loss. Each agent is optimised through Q-learning, and it has been shown that cooperative agents can solve complex trading problems~\cite{LEE_2007} 


Lee et al. \cite{LEE_2020} proposed a Multi-agent Portfolio Management System (MAPS) to capture the dynamic nature of financial markets. Each MAPS agent operates independently, managing its portfolio. The challenge of identifying the best joint policy is addressed by training each agent separately with DQN, sharing the state space and experience replay. The objective is to maximise rewards while ensuring diverse actions from each agent. The loss function is a weighted average of global loss and local losses (individual agent losses), reflecting the goal of maximising rewards and promoting diversity among agents. This inspires further exploration, including investigating changes in agents' utility functions relative to risk aversion, adjusting agent weighting based on market conditions, and applying transfer learning to refine approaches by limiting the information each agent receives.

AbdelKawy et al. \cite{ABDELKAWY_2021} introduced a synchronous-based multi-agent DQN/DDPG trading model. Each stock was initially trained with a DQN agent using a ResNet \cite{HE_2016} network architecture, followed by transfer learning applied to each trading module to reduce computational time. Shavandi and Khedmati \cite{SHAVANDI_2022} developed the first multi-frame DQN multi-agent system with three agents operating at different time frequencies (1 hour, 15 min, and 5 min). The lowest-frequency agent's output serves as input for higher-frequency agents, reflecting the fractal nature of financial markets \cite{PETERS_1994}. While the combined agent outperformed individual components, this was based on a single foreign exchange pair (EUR/USD), warranting cautious interpretation. Other studies have also explored various aspects of multi-agent learning in financial markets, including work by Casgrain et al. \cite{CASGRAIN_2022}.
\subsection{Model Interpretability\label{subsec:Model-Interpretability}}

Wang et al. \cite{WANG_2019} developed an interpretable investment strategy to answer \textquotedblleft What kind of stocks are selected as winners?\textquotedblright{} They used sensitivity analysis methods \cite{ADEBAYO_2018, WANG_2016_B, WANG_2018} to examine input factors' influence on final outputs. Their results showed that the algorithm selects stocks with favourable scores in long-term growth, low volatility, high intrinsic value, and recent undervaluation. Cong et al. \cite{CONG_2020} used a similar framework with polynomial sensitivity analysis. Despite the importance of model interpretability, explainability, and causality in financial research, the current RL literature has not addressed these themes extensively. Only Wang et al. \cite{WANG_2019} and Cong et al. \cite{CONG_2020} consider model interpretability, while Wang et al. \cite{WANG_2021_A} highlight the benefits of identifying causal structures among stocks using the PC \cite{SPRITES_2001}. They then employ this output as an input for graph convolution networks.

For RL in QF, model interpretability, explainability, and causality are crucial to building trust and ensuring that investment strategies are transparent and comprehensible to stakeholders. Investment managers must articulate the performance of the strategy to stakeholders. Without explanatory capacity, the adoption of the strategy may be limited. 
Incorporating causal inference methods to interpret relationships between input features and RL actions could be a promising research avenue \cite{PETERS_2017}.

\section{RL Applications in Modern Finance -- Promising Areas\label{sec:RL-Applications-in}}


This section reviews common modern finance applications and RL-based solutions and examines RL framework customisation for classic QF issues. The literature illustrates RL's adaptability to the financial domain as can be seen in the range of popular applications listed in Table \ref{tab:ThemsAndApplications} (in Appendix). 
In the following, we provide an examination of the salient ideas and potential extensions within the current literature on this topic and provide future directions in these areas. In addition, we delineate the pivotal concepts and thematic trends that have emerged. Although we acknowledge the extensive literature on PM, we have reserved some works for inclusion in different subsections as they emphasise distinct methods or themes relevant to PM. For example, the papers by Ding et al. \cite{DING_2018} and Yu et al. \cite{YU_2019}, which use IL methods within the PM environment, are already discussed in subsection \ref{subsec:Imitation learning}.


\subsection{Portfolio Management Under the RL Framework\label{subsec:PortManagement}}

Markowitz's seminal work on portfolio theory \cite{MARKOWITZ_1952} remains a foundational text in investment management, inspiring numerous contemporary Portfolio Management (PM) strategies. However, its utility relies on assumptions that, upon scrutiny, may seem overly simplistic. Among these assumptions are the rational behaviour of investors and the efficient operation of markets, both of which are often contradicted by empirical evidence. In particular, investors frequently exhibit herd-like behaviour towards speculative investments, culminating in asset bubbles or market crashes due to large-scale financial asset sell-offs \cite{LUX_1995, BARBERIS_2003}. Against this backdrop, RL-based PM frameworks, devoid of these stringent assumptions, have attracted considerable research interest. Their promising results indicate the potential for superior performance compared to the classical Markowitz framework.

Initial applications of RL in QF, particularly PM, were scarce. Jagmin et al. \cite{JANGMIN_2006} presented an RL agent using a Value-based-framework capable of dynamically allocating strategies across risky assets. The agent used four distinct trading patterns to predict share prices, followed by a meta-policy to distribute investments among these patterns optimally. However, the Value-based model's fundamental drawback was its dependence on a discrete action space, which misaligned with portfolio managers' preference for continuous weight allocation. Park et al. \cite{PARK_2020} applied the Value-based framework to an Exchange-Traded Funds (ETFs) portfolio, highlighting the potential shortcomings of a discrete action space and suggesting potential solutions. However, it should be noted that a discrete action space still lacks the necessary realism.

Arguably, the first attempt to develop a more realistic PM framework was made by Jiang et al. \cite{JIANG_2017,JIANG_2017_1}. They designed a model to manage a diverse cryptocurrency portfolio, with the agent aiming to maximise the average cumulative return $R$ across 12 cryptocurrencies (with Bitcoin as a cash proxy) at a given time $t$. Despite its pioneering nature, Jiang et al.'s work leaves room for enhancements, including the incorporation of constraints in the optimisation. In response, building upon their earlier work \cite{ALMAHDI_2017}, Almahdi and Yang \cite{ALMAHDI_2019} proposed the first constraint optimisation problem within the RRL framework. Their application of Particle Swarm Optimisation (PSO) \cite{KENNEDY_1995} to incorporate cardinality, quantity, and round lot constraints \cite{LIAGKOURAS_2018} transformed the unconstrained RRL case into a constrained optimisation problem, effectively combining the RL and PSO frameworks. However, directly incorporating constraints within the RL framework might offer additional benefits. In a further advancement, Wang \cite{WANG_2019_B} and Wang and Zhou \cite{WANG_2020} proposed a continuous-time mean-variance portfolio framework with entropy regularisation to balance exploration and exploitation. Named Exploratory Mean-Variance (EMV), this framework includes policy evaluation, policy improvement, and a self-correcting mechanism for Lagrange multipliers. By avoiding deep structures, the authors sought to bypass issues related to low interpretability, extensive hyperparameter tuning, and unstable performance.

\subsection{Financial Options-Based RL Applications}

The evolution of option\footnote{An \textit{option} is a financial contract where the owner has the right, but not the obligation to exercise, to buy or sell the reference instrument at a predefined strike.} pricing methodologies, stemming from seminal works by Black and Scholes \cite{BLACK_1973} and Merton \cite{MERTON_1973}, has spawned extensive literature. These foundational papers, collectively known as the Black-Scholes-Merton (BSM) model\footnote{The BSM model estimates option pricing by considering factors like asset price and volatility, estimating the option premium based on contract execution probability.}, assume complete markets \cite{ARROW_1954}, where every financial instrument can be replicated without transaction costs. However, market frictions, such as transaction costs and liquidity constraints, violate this hypothesis. Furthermore, the BSM model presumes constant volatility, often an inaccurate assumption. RL, unbound by these simplified assumptions, provides a robust alternative.

The hedging process typically involves trading in the underlying asset of the option, such as stocks, as dictated by the first derivative of the option pricing formula \cite{BLACK_1973, MERTON_1973}. The majority of works in this domain explore optimal hedging amidst transaction costs in markets that do not adhere to the completeness principle, and there are differences between the European and American options. 
While the classic European option pricing employs the BSM model, the American option pricing requires a more intricate approach, relying on the Least-Squares Monte Carlo (LSMC) algorithm \cite{LONGSTAFF_2001}.

Li et al. \cite{LI_2009} pioneered the pricing of American options using RL, using the Least-Squares Policy Iteration algorithm (LSPI) \cite{LAGOUDAKIS_2003} and a variant of Q-learning, namely Fitted Q-learning \cite{GORDON_1995}, for learning exercise policies for American options. These methods were demonstrated to outperform the LSMC benchmark for both simulated and real data. Subsequently, Dubrov \cite{DUBROV_2015} expanded this framework to price Convertible bonds—corporate bonds that can be converted into the issuer's equity. Furthermore, a non-RL model using Random Forests was proposed, outperforming the RL-based benchmarks (LSPI and Fitted Q-learning) and the standard LSMC. However, no recent studies have applied RL to American option pricing.

Buehler et al. \cite{BUEHLER_2019} applied RL to optimal hedging with transaction costs, testing their model on vanilla and barrier options. Similarly, Vittori et al. \cite{VITTORI_2020} applied Trust Region Volatility Optimisation \cite{BISI_2019}, incorporating features such as Strike, Call price, and Delta hedge\footnote{Delta hedging in options is a strategy that aims to reduce the directional risk associated with the underlying asset's price movements.}, as well as the preceding action in their environment. Conversely, Kolm and Ritter \cite{KOLM_2020} omitted the Delta hedge, arguing that it is a non-linear function of state variables. Therefore, the agent should be able to identify such patterns. Not including such functional forms avoids relying on BSM assumptions. Du et al. \cite{DU_2020} proposed a similar strategy with Deep Q-learning and Pop-Art \cite{vanHASSELT_2016}. Halperin \cite{HALPERIN_2020} developed a discrete option pricing model using Q-learning, aiming to determine both the option price and the hedging strategy, without considering transaction costs. Cannelli et al. \cite{CANNELLI_2020} demonstrated that a risk-averse contextual k-armed bandit is superior to DQL in sample-efficiency and hedging error reduction under the BSM framework. Lastly, Cao et al. \cite{CAO_2021} redefined the objective function of Almgren and Criss \cite{ALMGREN_2001} (AC)\footnote{Almgren and Criss \cite{ALMGREN_2001} proposed a quadratic utility as an objective function for the optimal execution problem. This transforms the optimisation problem into a mean-variance problem, with the expected hedging cost and the variance of the hedging cost being the two key components of the objective function.}, defining it as follows:
\begin{equation}
Y\left(t\right)=\mathbb{E}\left(C_{t}\right)+c\sqrt{\mathbb{E}\left(C_{t}^{2}\right)-\mathbb{E}\left(C_{t}\right)^{2}},\label{eq:-36-1}
\end{equation}
where $c$ is a constant, and $C_{t}$ represents the total hedging cost from time $t$ onwards. The objective is to minimise $Y\left(0\right)$. They subsequently fitted two separate Q-functions to track the expected value of the transaction cost $\mathbb{E}\left(C_{t}\right)$ and the expected value of the square of the cost $\mathbb{E}\left(C_{t}^{2}\right)$. As noted by the authors, this methodology offers various benefits over the baseline, such as the broader range of objective functions and the learning algorithm that supports continuous state and action space.

These applications represent intriguing attempts into option hedging and further testify to the adaptability of the RL framework. A compelling extension would be to apply such methodologies to more complex instruments, such as exotic options.

\subsection{Order Execution\label{subsec:Order-execution}}

In the trading ecosystem, an asset manager, acting as a client, delegates a portfolio of trades, typically to a broker. This broker executes these trades, either selling or buying a predefined volume, with the central objective being transaction cost minimisation through strategic distribution across varied timeframes. Cartea et al. \cite{CARTEA_2015} highlight a key challenge - balancing market impact mitigation and price risk. Much of the existing RL literature in QF assumes negligible market impact, given the relatively small position size. Classic financial references such as Bertsimas and Lo \cite{BERTSIMAS_1998} and AC offer valuable insight, but their assumptions are often too strict and usually ignore the dynamic interplay that defines financial markets, offering a relatively static framework. 

The liquidity of the underlying stocks considerably influences the trading process, making the trading of less liquid stocks more difficult because of potentially higher transaction costs. The market impact of large block trades represents another substantial challenge. The current RL literature often presupposes immediate execution of trading positions, an assumption that may not accurately reflect real-world scenarios. Only a few researchers, such as Wang et al. \cite{WANG_2021_B}, incorporate this aspect into their RL frameworks, evidenced by their inclusion of an execution module in a Hierarchical RL-based framework.

The large-scale implementation of RL in trade execution was pioneered by Nevmyvaka et al. \cite{NEVMYVAKA_2006}, who used a modified Q-learning agent with 1.5 years of millisecond LOB data from NASDAQ. They classified state representation variables into private and market variables, a tactic subsequently adopted~\cite{LIN_2020} and expanded by incorporating a risk-averse framework \cite{SHEN_2014} and volume/spread attributes \cite{HENDRICKS_2014}. Bao and Liu \cite{BAO_2019} proposed the first multi-agent application of RL to optimal liquidation. They expanded the AC framework to the multi-agent environment, examining the interaction of cooperative and competitive behaviours and their implications for the overall system. Later, Bao \cite{BAO_2019_B} addressed the issue of fairness \cite{MOULIN_2004} in the multi-agent framework, exploring inequities that arise when differential execution strategies are applied to trades of the same assets with varying quantities and time horizons. Bao suggested adjusting each agent's reward by referencing it against the classic Gini or Bonferroni indices \cite{WEYMARK_1981}. Despite their results being based on simulated data and a limited number of agents, the concept of fairness remains deeply relevant, not just to trade execution but to various facets of the financial world as well which remains as an open research area. Fang et al. \cite{FANG_2021} introduced a PPO-based framework with PD to handle imperfect information, integrating a quadratic penalty for market impact, following Ning et al. \cite{NING_2020} and Lin and Beling \cite{LIN_2020}. Other notable studies include \cite{DABERIUS_2019, KARPE_2020, NING_2020}.



Overall, although the AC model offers a convenient closed-form solution for order execution, its critical assumptions can lead to suboptimal outcomes when applied to real-world scenarios. In contrast, RL-based methodologies provide a more dynamic and flexible framework for trade execution, allowing for the adaptation to market conditions in real-time. This adaptability is crucial for managing the complexities of market impact and transaction costs, making RL a potentially more effective approach in the evolving landscape of order execution.

\subsection{Market-Making\label{subsec:Market-making-and}}

A market maker's function is to ensure liquidity for buyers and sellers by persistently offering bid and ask quotes alongside the respective market sizes. The profit avenue for the market maker is the spread, the difference between the bid and the ask. One substantial risk market makers face is the inventory risk, which arises when the current inventory remains unsold. This can occur when informed traders engage with market makers prior to a price drop, resulting in losses. Therefore, maintaining minimal inventory is a key goal for market makers, achievable either by initiating trades to reduce the inventory at a cost or by skewing asset prices to attract trades offsetting the inventory.

With the increase in trading frequencies over recent decades, human processing of the resulting data flood has become nearly impossible. This underscores the critical importance of electronic Limit Order Book (LOB). Given this, applications in market-making have the potential to attract significant interest. However, most current applications are based on simulated or artificial data, and few use real LOB data. However, there is a growing trend to use real-world data, as exemplified by Zhao and Linetsky's 2021 study using LOB data from the Chicago Mercantile Exchange (CME) on S\&P 500 and 10-year Treasury note \cite{ZHAO_2021}.

Chan and Shelton's pioneering work \cite{CHAN_2001} marked the earliest application of RL to market-making. They proposed the first electronic market maker to quote bid and ask prices, drawing inspiration from Glosten and Milgrom's seminal work \cite{GLOSTEN_1985}. Following this, the field has seen notable advancements with the demonstration by Spooner et al. \cite{SPOONER_2018} that an asymmetrically dampened reward function improved learning stability, and Ganesh et al. \cite{GANESH_2019} constructed a competitive multi-agent based on PPO. In addition, Guéant and Manziuk \cite{GUEANT_2019} proposed the first market-making framework for the optimisation of multiple corporate bonds under a model-based Actor-Critic RL framework, marking a significant step in the field.

Avellaneda and Stoikov's classic market-making model \cite{AVELLANEDA_2008} inspired Spooner and Savani \cite{SPOONER_2020} to introduce a game-theoretic approach to market-making using adversarial RL. Gašperov and Kostanjčar \cite{GASPEROV_2021} proposed a derivative-free adversarial neuroevolution-based RL \cite{SUCH_2017} market-making agent. Zhao and Linetsky \cite{ZHAO_2021} included the Book Exhaustion Rate (BER) as part of the features to protect market makers from informed traders. The most recent publication by Zhong et al. \cite{ZHONG_2021} used a Q-learning framework to create a market-making agent with the aim of maximising the expected net profit. Their approach outperformed benchmark strategies using historical LOB data, garnering the interest of a company intending to implement this framework. However, more real-world implementation of such strategies and thorough examination of their effectiveness are necessary for future research.


\subsection{End-to-End Trading Systems\label{subsec:EndToEnd}}

Within the surveyed literature, most publications address a single facet of quantitative trading, such as PM or execution. However, a handful of studies have sought to incorporate multiple elements of trading, creating intriguing end-to-end systems. These systems showcase the adaptability of the RL framework across various aspects of QF, representing an initial step towards an autonomous trading system that minimises human intervention.

Lee and Jagmin's work \cite{LEE_2002} is notable for being one of the first attempts at creating such a system. In their research, two agents generate trading signals while the remaining two manage order placement (for more details, see \ref{subsec:Multi-agent-applications}). In a subsequent study, Dempster and Leemans \cite{DEMPSTER_2006} developed an RL trader that integrates with a broader trading system that emphasises risk management. The proposed system, designed for FX markets, is based on three pillars: an ML algorithm \cite{MOODY_1997}, risk and performance management, and utility-driven dynamic optimisation. The authors introduced several adjustments to the primary concept, using \textquotedbl one-at-a-time random search optimisation\textquotedbl{} to navigate the system's inherent complexity.

In recent efforts, Patel et al. \cite{PATEL_2018} proposed a collaborative multi-agent market-making system that uses DQN. Here, the \textquotedbl Macro agent\textquotedbl{} operates on minute tick data to decide whether to buy, sell or hold an asset. Subsequently, a \textquotedbl Micro agent\textquotedbl{} leverages the order book data to determine the order placement within the LOB. Similarly, Wang et al. \cite{WANG_2021_B} addressed the common assumption in the existing literature that each portfolio allocation can be executed instantaneously, thus overlooking price slippage. They introduced a Hierarchical Reinforced Framework for Portfolio Management (HRPM), incorporating high-level portfolio management and low-level trading execution policies. The former is optimised through REINFORCE with an additional entropy term to promote diversification, while the latter uses the Branching Dueling Q-Network \cite{TAVAKOLI_2018}.

These approaches of Patel et al. \cite{PATEL_2018} and Wang et al. \cite{WANG_2021_B}, which combine crucial elements of QF into a single trading system, serve as the first significant advancements since the groundbreaking work of Mnih et al. \cite{MNIH_2013,MNIH_2015}. One possible enhancement to these approaches could be the integration of a risk management module \cite{DEMPSTER_2006}.

\subsection{Business Cycles and Augmenting Existing Trading Strategies}

RL has emerged as a powerful tool in financial trading, with its ability to learn and optimise from iterative interactions in a dynamic environment. In this subsection, we examine the innovative use of RL in the tuning and optimisation of trading strategies.

Inspired by seminal work on business cycles \cite{KITCHIN_1923}, Tan et al. \cite{TAN_2011} developed a trading algorithm that capitalises on stock price cycles. The core idea of their approach is to initiate and terminate trades as close as possible to each stock's turning point. This strategy first involves identifying asset-specific cycles optimised via RL within a Value-based framework. The authors proceeded to design a bespoke reward function suited to their requirements. Turning points, which signal when to open or close a position, were identified through an Adaptive Network Fuzzy Inference System. The final component is the trading agent, whose goal is to execute trades in close proximity to these determined turning points, once again harnessing RL within the Value-based framework for identification. The unique feature of this application is the deployment of RL as a constituent part of a more comprehensive trading strategy, underscoring its potential for fine-tuning various system components. Eilers et al. \cite{EILERS_2014} provide another example of RL's application in refining trading strategy parameters by leveraging recognised seasonal effects \cite{HANSEN_2005, ARIEL_1987, FRENCH_1980}. Their RL-based strategy evaluates whether to generate long/short signals, holding periods, and leverage levels before event days.

Kim and Kim \cite{KIM_2019} implemented a DQN for pair trading, rewarding the agent when the spread between two similar companies reaches a predefined threshold and reverts to the mean. The agent is penalised if stop-loss limits are hit or the spread fails to revert. The environment focuses solely on the spread but could be enhanced by incorporating factors like liquidity. Wang et al. \cite{WANG_C_2021} applied a similar strategy to five-year data from 75 NASDAQ companies, outperforming the baseline in terms of SR. Cartea et al. \cite{CARTEA_2021} are perhaps the first to apply DDQN and Reinforced Deep Markov Models (RDMMs) \cite{FERREIRA_2020} to FX triplets. In an FX triplet, one of the three FX pairs is redundant due to the no-arbitrage rule in frictionless markets. Therefore, establishing an opposing position is expected to generate profit when the triplet deviates from the no-arbitrage relationship. Through a simulation study, the authors favoured the use of RDMMs. Kim et al. \cite{KIM_2022} proposed a hybrid RL framework in which the first component, based on the TD3 algorithm (Actor-Critic) \cite{FUJIMOTO_2018}, is responsible for trading actions in pairs trading, while the second component, based on a DDQN agent, is responsible for stop-loss boundaries. In recognition of the risk of a structural break in the trading pair, Lu et al. \cite{LU_2022} introduced a hybrid RL-based on DQN in which structural break detection mechanisms are used as input to a DQN agent.


In summary, these studies illustrate the versatility and robustness of RL in enhancing trading strategies, demonstrating its ability to adapt to complex market dynamics and optimise execution in response to changes. Future research might explore the applicability of these methodologies, especially in markets with different characteristics or diverse economic conditions.

\subsection{Asset Allocation}

The versatility of the RL framework has been evidenced through various applications, including parameter tuning, exemplified by Tan et al. \cite{TAN_2011}, who deployed RL to optimise parameters to discern stock-specific cycles. However, the potential role of the RL in strategic allocation - referring to long-term asset allocation - and market timing, which involves switching capital between asset classes or financial markets based on predictive methods, is not yet thoroughly explored in existing research. A notable exception is the work of Hens and Wöhrmann \cite{HENS_2007}, who apply the RRL framework \cite{MOODY_1998} for the dynamic allocation between bonds and equity indices. The allocation between the stock and bond indices is expressed as the portfolio return $R_{t}^{P}\in\mathbb{R}$, as follows:
\begin{equation}
R_{t}^{P} = w_{t}^{\theta}R_{t}^{B} + \left(1-w_{t}^{\theta}\right)R_{t}^{E},\label{eq:-36}
\end{equation}
Here, $R_{t}^{B}$ and $R_{t}^{E}$ denote bond and equity returns at time $t$ respectively, and $w_{t}\in\left[0,1\right]$ represents weights under an exponential parametric function based on $\theta$. 
Similarly, Du et al. \cite{DU_2009} switch allocations between risk-free and risky assets and
a Value-based approach \cite{PENDHARKAR_2018} was used to allocate between the S\&P500 ETF and the AGG Bond Index or the 10-year US Treasury note.

Although the described framework appears simplistic, it uncovers novel avenues for applying RL in finance. The Hens and Wöhrmann framework \cite{HENS_2007} is a valuable tool for fund managers who frequently need to alternate between asset classes. However, there are opportunities to extend their work by incorporating more than two asset classes and integrating exogenous macroeconomic variables into the environment. Given that fund managers often consider macroeconomic conditions when strategising capital deployment for their clients, such enhancements could increase the relevance and application of this approach in the real world.

\subsection{Human in the Loop -- Robo-Advising\label{subsec:HumanInTheLoop}} 
Most applications discussed aim to reduce human intervention by implementing autonomous trading systems, with Robo-advising as a notable exception where human input is crucial. Defined as algorithms that cater to client-specific risk appetites, Robo-advisors offer a cost-effective alternative to traditional investment management, minimising overheads and human biases in asset allocation and recommendations \cite{FOERSTER_2017_A}. Historically, risk preferences were gauged through potentially biased questionnaires \cite{CHANRNESS_2013, HOLT_2002}, whereas contemporary approaches like Inverse Reinforcement Learning (IRL) and Inverse Portfolio Optimisation (IPO) dynamically estimate these preferences \cite{ALSABAH_2021, WANG_2021_D, YU_2020}.

The reviewed studies highlight that Robo-advisors iteratively learn an investor's risk tolerance by adjusting portfolio allocations based on the investor's feedback. This interactive process, however, introduces a trade-off between exploiting known preferences and exploring potentially more optimal allocations, adding complexity due to the variability of investor preferences over time \cite{ALSABAH_2021}. Innovations include a dual-agent framework where one agent infers preferences and expected returns, while the other optimises investments over multiple periods \cite{WANG_2021_D}. Further applications in wealth management use advanced algorithms like G-Learning to refine these methods \cite{DIXON_2020}. Despite Robo-advising's growing industry significance, it remains under-researched. Future studies could focus on multi-agent systems in Robo-advising to understand competitive or cooperative dynamics in risk preference management, potentially revealing broader implications for investor behaviour.

\section{Discussion\label{sec:Results-Discussion}}

In this survey of 167 research articles, we discovered numerous compelling concepts for practical trading applications. RL agents generally exceeded their respective benchmarks, with notable exceptions such as Sherstov and Stone \cite{SHERSTOV_2004}. The robustness of these frameworks is manifested across various global regions and asset classes. For example, Maringer and Ramtohul \cite{MARINGER_2010} examined the performance of equity indices in the UK, France, Germany, and Switzerland, while Bekiros \cite{BEKIROS_2010} focused on stock indices in Japan and the United States. Bisht and Kumar \cite{BISHT_2020} honed in on Indian stock indices, with others, including Sawhney et al. \cite{SAWHNEY_2021} and Fang et al. \cite{FANG_2021}, extending their frameworks to the Chinese and Hong Kong financial markets. The application of the RL framework to diverse asset classes is presented in the surveyed literature. Carapuço et al. \cite{CARAPUCO_2018} and Sornmayura \cite{SORMAYURA_2019} explored different currencies, Deng et al. \cite{DENG_2016} investigated commodities, and Gueant and Manziuk \cite{GUEANT_2019} delved into fixed income. Cryptocurrencies such as Bitcoin and Ethereum were the focus of Sattarov et al. \cite{SATTAROV}, while Ye et al. \cite{YE_2020} studied a combination of cryptocurrencies and stocks. Further extensions to multiple assets using different RL algorithms were made by authors such as Zhang et al. \cite{ZHANG_2020_A}. Yuan et al. \cite{YUAN_2020} and Katongo and Bhattacharyya \cite{KATONGO_2021} performed comparative analyses across three main Actor-Critic algorithms. Notable contributions were also made by Lavko et al. \cite{LAVKO_2023}, who evaluated the trading performance of various model-free agents, comparing them with the classical mean-variance framework \cite{MARKOWITZ_1952} and equally weighted portfolios. RL agents consistently surpassed the respective benchmarks in these comparisons.

Despite the impressive performance of RL agents across regions and asset classes, concerns arise about the evaluation practices in the literature. Benhamou \cite{BENHAMOU_2020_A} compares an RL PM framework with classic strategies \cite{MARKOWITZ_1952}. However, this comparison may be unbalanced, as the RL framework uses a different information set than the benchmarks. Hence, a comparison with the basic framework \cite{MARKOWITZ_1952} could be incomplete. A fairer comparison with all models using identical input data could be ideal, thereby offering a genuine reflection of each method's data-leveraging ability and providing a more authentic evaluation of their merits.

Numerous studies use the SR as a performance metric. However, excessively high SR values warrant careful scrutiny. Since it can be a product, for example, of overfitting \cite{BAILEY_2014}  and data snooping bias \cite{SULLIVAN_1999}. Moreover, empirical evidence suggests that an SR above 3 is exceptional. Thus, findings such as Deng et al. \cite{DENG_2016} with an SR of 21.2, or Xu et al. \cite{XU_2021} astonishing SR of 217.68, raise questions. A possible explanation could be the forward-looking bias \cite{BENHAMOU_2020_A,BENHAMOU_2020_B}, as the authors typically observe prices at time $t$ and actions are taken at the same time. Introducing a lag could make these models more realistic, although more challenging.

In subsection \ref{subsec:Model-Interpretability}, we highlighted the rare instances of model interpretability in the current literature. Although Markowitz's methodology \cite{MARKOWITZ_1952} is simple, its transparency aids in understanding the portfolio weight output, contrasting with the complex frameworks often found in recent RL literature. For example, Gold \cite{GOLD_2003} discovered that lagged returns were more significant than the count of hidden neurons, although these results varied by currency pair. This finding revealed complexities, as the optimal parameters varied by currency and exhibited interdependence. From a practical viewpoint, it is crucial to have a robust rationale for parameter selection and a clear explanation for any observed variances. When introducing a new algorithm into a trading system, the three primary considerations are the complexity it imposes, the explicability of its performance, and the investment rationale. Failing to justify these aspects can erode transparency and possibly lead to misrepresentation of investment products. In the reviewed literature, attempts to justify complex methodologies are sparse. Exceptions include Ponomarev et al. \cite{PONOMAREV_2019} grid-search strategy to optimise NN architecture. Recently, Aboussalah and Lee \cite{ABOUSSALAH_2020} proposed an RRL-based agent capable of hyper-parameter tuning via Bayesian optimisation. This automated Gaussian process, with Expected Improvement as the acquisition function, enables the trading system to select architectures optimised for maximising the selected reward.

There is a compelling need to expand the existing literature to incorporate a detailed examination of trading strategies based on conventional factors, especially in the context of PM, as introduced in subsection \ref{subsec:Fundamental-data-and}. These stylised investment factors, backed by a robust body of literature, call for a comprehensive understanding of their influence on the performance of an RL trading system. A truly effective alpha strategy\footnote{An investment strategy that aims to generate returns that exceed the performance of a benchmark index, after adjusting for systematic (market) risk.} based on RL should exhibit significance after accounting for those factors \cite{FAMA_1992, FAMA_1993, FAMA_1996}. Only recently, Cong et al. \cite{CONG_2020} evaluated their RL agent using a framework similar to the one aforementioned. They convincingly demonstrated the robustness of their strategy even after integrating several control variables.

One critique, inextricably linked to the preceding discourse, relates to the trend of treating RL in QF predominantly as an \textquotedblleft engineering \textquotedblright{} issue. This is evident from the tendency of authors to draw heavily on ML and DL literature, applying these concepts to QF after making minor modifications or exploring alternative NN architectures and feature selection mechanisms. This approach can be enriched by incorporating conceptual and innovative developments, as advocated in Section \ref{sec:ML-Themes-and}. Future research should look beyond engineering solutions, striving for innovation relevant to the QF domain, as seen in seminal works by Moody and Wu \cite{MOODY_1997}. The overengineering of solutions brings forth another concern – the vast array of alternative solutions based on varied network architectures makes it challenging to identify superior approaches and pinpoint the essential components within a given solution.

Another issue to address is the evident survivorship bias in some publications, for example, Kaur \cite{KAUR_2017} and Wu et al. \cite{WU_2020}. These works often apply their frameworks to successful single stock names like Apple Inc., neglecting to test their methodology on stocks that are currently non-existent or bankrupt. Furthermore, the focus on a limited set of tradable assets introduces potential survivorship bias, and the success of these assets does not necessarily imply effectiveness across other available assets. To lend greater authenticity, we should scrutinise the proposed trading system on benchmark constituents over time. Recent studies have shown progress in this regard. For example, Lee et al. \cite{LEE_2020} extended their RL framework to the Russell 3000, while Wang et al. \cite{WANG_2019} used Wharton Research Data Services (WRDS) to select a quarter of valid stocks annually.

\section{Future Directions\label{sec:Future-Aseas-for-Research}}


\noindent \textbf{Exploration of Alternative Features: }In subsection \ref{subsec:Features}, we underscore the potential for investigating features derived from alternative data sources, such as ESG and macroeconomic-based features. Accurate environment modelling is crucial for successful RL, especially in the chaotic world of financial markets. Consequently, finding meaningful features can determine the success or failure of a trading system. Although exploring alternative data could present challenges like data retrieval and time-consuming preprocessing, the potential benefits merit further investigation.

\vspace{1.5mm}
\noindent\textbf{Knowledge transfer in trading systems: }Subsection \ref{subsec:Transfer-learning} highlights the potential for applying transfer learning in financial markets. This method could expedite the training process and complete missing historical data for stocks with a brief history. The limited available literature \cite{WEI_2019} points to the potential of combining model-based RL with transfer learning, as it could lead to enhanced performance and faster convergence by reducing data dependency. 

Meta-learning \cite{SCHMIDHUBER_1987,THURN_1995}, as a way of generalising knowledge from multiple tasks, presents another exciting research opportunity. This method could allow the agent to learn new tasks faster and better, while providing a natural regularisation method to prevent overfitting and enhance generalisation performance. A promising application of meta-learning in financial trading systems could be its use when an agent needs to adapt to a new market or set of assets. Moreover, multi-task learning—a related concept in which a single network is simultaneously trained on multiple related tasks—presents a compelling approach to accelerate the training process and improve regularisation \cite{CARUANA_1998}. Its potential is especially evident in contexts such as training on stocks within the same sector or analysing LOB data of similar stocks.

\vspace{1.2mm}
\noindent \textbf{Multi-agent solutions: }Subsection \ref{subsec:Multi-agent-applications} examined multi-agent applications and identified future research directions. Recognising that financial markets are essentially MAS, their exploration is crucial for revealing hidden market structures and enhancing our understanding of these intricate systems. A key challenge is scalability: the existing literature often restricts its focus to a few agents, pointing to the need for wider research. MAS also present an opportunity to assess the market impact of individual agents, challenging the assumption of an unaffected environment. Another underexplored aspect is the establishment of coordination and communication protocols among different agents in the QF context \cite{FOERSTER_2016}. Effective protocols can reduce information asymmetry, increase market transparency, and improve market efficiency. For instance, within a single firm, if one trading agent uncovers a flaw in a common strategy, this discovery could be shared quickly, allowing for strategic adjustments, improved overall performance, and minimised capital losses. Hence, this aspect warrants further examination. Finally, we eagerly anticipate further research exploring integrating MAS with various ML concepts and financial applications, as underscored in subsection \ref{subsec:HumanInTheLoop}.

\vspace{1.5mm}
\noindent \textbf{Multi-objective RL: }In subsection \ref{subsec:PortManagement}, we detailed RL's relevance in PM. Future research could target a more natural constraint of the action space, bypassing heuristic methods. This includes constraints at various levels such as sectors, countries, and individual stocks. This prompts a demand for a multi-objective RL framework addressing conflicting objectives (e.g., \cite{MOFFAERT_2014}). We might draw from Markowitz's PM approach \cite{MARKOWITZ_1952}, which balances conflicting objectives through a convex optimisation problem, enabling a tradeoff between returns and variance. Therefore, any model aspiring to supersede Markowitz's must offer superior performance and comparable flexibility. This observation logically moves us towards the promising idea of multi-objective RL in finance.

\vspace{1.5mm}
\noindent \textbf{Different applications of Robo-adviser: }In subsection \ref{subsec:HumanInTheLoop}, we examined Robo-advisers in PM and suggested future research. Human-in-the-loop methodology exploration should expand beyond this. An intriguing trajectory, outlined by Alsabah et al.~\cite{ALSABAH_2021}, is Robo-advisers for tax loss harvesting\footnote{Tax loss harvesting refers to offsetting capital gains tax liabilities on appreciated investments by selling those that have experienced a loss.} \cite{STEIN_1999}, offering the potential to decrease temporal and monetary expenditure.

\vspace{1.5mm}
\noindent \textbf{Hierarchical RL: }While Hierarchical RL (HRL) is a concept of significant potential within RL literature~\cite{LEVY_2017, LI_2019_B}, it is noticeably underexplored in QF apart from a few exceptions~\cite{WANG_2021_B}. HRL leverages hierarchical relationships among agents, breaking down complex problems into smaller tasks, thus enabling faster resolutions. In QF, this could streamline processes like asset selection and execution or further subdivide asset selection into a hierarchy of steps for more efficient problem-solving.

\vspace{1.5mm}
\noindent \textbf{Single vs. many stock trading systems: }Trading systems typically target single assets or multiple assets. Future research might construct portfolios from several single-asset trading systems. This would harness the idiosyncratic detail of single-asset systems and the holistic perspective of multi-asset systems, underpinning a comprehensive asset management strategy \cite{ABOUSSALAH_2022}.

\vspace{1.5mm}
\noindent \textbf{Human in the loop: }In subsection \ref{subsec:HumanInTheLoop}, we discussed instances requiring human feedback as an integral part of the RL framework. Another potential application arises within the context of PD or IL. Notably, when the expert or teacher lacks access to an optimal solution but has a preferred strategy, the process of distillation or IL could aim to replicate this. This approach could be especially valuable when the goal is to integrate human expertise or specific strategic intentions into the automated decision-making process.

\section{Conclusion\label{sec:Conclusion}}

We conducted a comprehensive survey of 167 publications exploring RL applications in QF. Our findings reveal a promising alternative to classic QF methodologies, particularly in areas such as PM and option hedging. We also examined the practical use of ML concepts, including transfer learning, imitation learning, and multi-agent RL, underscoring their potential impact on future research. Throughout our analysis, we emphasise the fundamental components of RL: environment, rewards, and action. We discussed key advancements in these areas from a QF viewpoint, scrutinising their contribution to RL development in QF.

Although acknowledging the progress made in recent publications, we also identified challenges and limitations in earlier stages of the literature, particularly with respect to results and methodologies. As QF is an inherently applied field, it is crucial that the proposed solutions align with real-world conditions and meet rigorous standards closely. Looking ahead, we put forward several ideas for future research in various sections of our survey. In conclusion, our survey provides valuable insights into the current landscape of RL in QF and presents a roadmap for future research in this rapidly evolving field. By addressing the identified challenges and pursuing these research directions, we can further enhance the effectiveness and applicability of RL methods in QF, ultimately advancing the understanding and practice of intelligent decision-making in financial markets.


\begin{acks}
ChatGPT was used to shorten and refine the text for clarity and grammar. We provided short text segments, reviewed the outputs for quality and accuracy, and did not use ChatGPT to introduce new references or ideas beyond those in the original input.
\end{acks}



\newpage

\section*{APPENDIX}

\begin{table}[ht]
\centering
\fontsize{8pt}{10pt}\selectfont
\begin{tabular}{|l|l|l|l|l|l|l|l|l|l|l|l|}
\hline 
\textbf{Publication} & \textbf{Data Freq.} & \textbf{Period} & \textbf{PH} & \textbf{TI} & \textbf{CP} & \textbf{ME} & \textbf{PV} & \textbf{CH} & \textbf{OB} & \textbf{Oth.} & \textbf{FM} \\
\hline 
Neuneier \cite{NEUNEIER_1996} & Daily & 1986-1996 & x &  & x & x & x &  &  &  & \tabularnewline
\hline 
Moody and Wu \cite{MOODY_1997} & Monthly/30-min. & Several & x &  & x &  &  &  &  &  & \tabularnewline
\hline 
Moody et al. \cite{MOODY_1998} & Monthly/30-min. & Several & x &  & x &  &  &  &  &  & \tabularnewline
\hline 
Moody and Safell \cite{MOODY_2001} & Monthly/30-min. & Several & x &  & x &  &  &  &  &  & \tabularnewline
\hline 
Dempster et al. \cite{DEMPSTER_2001} & 1-min. & 1994-2002 &  & x & x &  &  &  &  &  & x\tabularnewline
\hline 
Nevmyvaka et al. \cite{NEVMYVAKA_2006} & Milliseconds & 1.5 years of LOB & x &  & x &  &  &  & x & x & \tabularnewline
\hline 
Dempster and Leemans \cite{DEMPSTER_2006} & 1-min. & 2000-2002 & x & x & x &  &  &  &  &  & \tabularnewline
\hline 
Lee et al. \cite{LEE_2007} & Daily & 1999-2005 & x & x &  &  & x &  &  & x & \tabularnewline
\hline 
Deng et al. \cite{DENG_2016} & Tick-level & Several & x &  &  &  &  &  &  & x & x\tabularnewline
\hline 
Jiang and Liang \cite{JIANG_2017} & 30-min. & 2014-2017 & x & x & x &  &  &  &  &  & x\tabularnewline
\hline 
Jiang and Liang \cite{JIANG_2017_1} & 30-min. & 2014-2017 & x & x & x &  &  &  &  &  & x\tabularnewline
\hline 
Almahdi and Yang \cite{ALMAHDI_2017} & Weekly & 2011-2015 & x &  & x &  &  &  &  &  & \tabularnewline
\hline 
Liang et al. \cite{LIANG_2018} & Daily & 2015-2017 & x &  &  &  &  &  &  & x & x\tabularnewline
\hline 
Xiong et al. \cite{XIONG_2018} & Daily & 2009-2018 & x &  & x &  &  & x &  &  & x\tabularnewline
\hline 
Jeong and Kim \cite{JEONG_2019} & Daily & Several & x &  &  &  &  &  &  &  & x\tabularnewline
\hline 
\end{tabular}
\caption{Mapping of Features for Most Cited Publications in the Survey. PH - Price History, TI - Technical Indicators, CP - Current Position, ME - Macroeconomic, PV - Profit/Value, CH - Cash, OB - Order Book, Oth. - Others, FM - Feature Mechanism.}
\label{tab:Data_and_Features}
\end{table}

\begin{table*}[h]
\centering{}
\small 
\begin{tabularx}{\textwidth}{p{3.4cm}p{1.8cm}x{0.5cm}cx{1.8cm}x{0.8cm}x{0.8cm}cx{0.8cm}c}
\toprule 
\textbf{Publication} & \textbf{Asset} & \textbf{\# of Act.} & \textbf{Cont.} & \textbf{Act. Func.} & \textbf{P. B.} & \textbf{Perf. Ratios} & \textbf{Util. Based} & \textbf{Other} & \textbf{Tr. Costs} \\
\midrule
Neuneier \cite{NEUNEIER_1996} & FX/Equities & 2 &  &  & x &  &  &  & x\\
Moody and Wu \cite{MOODY_1997, MOODY_1998, MOODY_2001} & FX/Equities & 2 &  & Tanh/Softmax & x & x &  &  & x\\
Dempster et al. \cite{DEMPSTER_2001} & FX & 2, 3 &  &  & x &  &  &  & x\\
Nevmyvaka et al. \cite{NEVMYVAKA_2006} & Equities & 11 &  &  &  &  &  & x & x\\
Dempster and Leemans \cite{DEMPSTER_2006} & FX & 2 &  & Tanh &  &  & x &  & x\\
Lee et al. \cite{LEE_2007} & Equities & 2-7 &  &  & x &  &  & x & \\
Deng et al. \cite{DENG_2016} & Com./ Eq. Index & 3 &  & Tanh & x &  &  &  & x\\
Jiang and Liang \cite{JIANG_2017, JIANG_2017_1} & Crypto & 12 & x & Softmax & x &  &  &  & x\\
Almahdi and Yang \cite{ALMAHDI_2017} & Stock ETF's & 5 & x & Logsig/Softmax &  & x &  &  & x\\
Liang et al. \cite{LIANG_2018} & Equities & 5 & x &  & x &  &  &  & x\\
Xiong et al. \cite{XIONG_2018} & Equities & 3 &  &  & x &  &  &  & \\
Jeong and Kim \cite{JEONG_2019} & Equity Index & 3 &  &  & x &  &  &  & \\
\bottomrule 
\end{tabularx}
\caption{Action and Reward types for the most cited papers in this survey. Asset = Asset Class, \# of Act. = Number of Actions, Cont. = Continuous, Act. Func. = Activation Function, P. B. = Profit Based, Perf. Ratios = Performance Ratios, Util. Based = Utility-Based, Tr. Costs = Transaction Costs. \label{tab:Actions-and-Rewards}}
\end{table*}

\begin{table*}[h]
\centering{}\fontsize{8pt}{10pt}\selectfont
\begin{tabular}{|l|x{1.5cm}|x{1.5cm}|x{1.5cm}|x{1.5cm}|x{1.5cm}|}
\hline 
\textbf{Publication} & \multicolumn{5}{c|}{\textbf{Applications}} \\
\cline{2-6}  
 & \textbf{Trading Systems} & \textbf{Portfolio Management} & \textbf{Asset Allocation} & \textbf{End-to-End} & \textbf{Trade Execution} \\
\hline 
\hline 
Neuneier \cite{NEUNEIER_1996} & x &  & x &  &  \\
\hline 
Moody and Wu \cite{MOODY_1997} & x &  &  &  &  \\
\hline 
Moody et al. \cite{MOODY_1998} &  &  &  &  &  \\
\hline 
Moody and Safell \cite{MOODY_2001} &  &  &  &  &  \\
\hline 
Dempster et al. \cite{DEMPSTER_2001} & x &  &  &  &  \\
\hline 
Nevmyvaka et al. \cite{NEVMYVAKA_2006} &  &  &  &  & x \\
\hline 
Dempster and Leemans \cite{DEMPSTER_2006} & x &  &  & x &  \\
\hline 
Lee et al. \cite{LEE_2007} &  & x & x &  &  \\
\hline 
Deng et al. \cite{DENG_2016} & x &  &  &  &  \\
\hline 
Jiang and Liang \cite{JIANG_2017} &  & x & x &  &  \\
\hline 
Jiang and Liang \cite{JIANG_2017_1} &  &  &  &  &  \\
\hline 
Almahdi and Yang \cite{ALMAHDI_2017} &  & x & x &  &  \\
\hline 
Liang et al. \cite{LIANG_2018} &  & x & x &  &  \\
\hline 
Xiong et al. \cite{XIONG_2018} & x &  &  &  &  \\
\hline 
Jeong and Kim \cite{JEONG_2019} & x &  &  &  &  \\
\hline 
\end{tabular}
\caption{Key application categories for the top 15 cited publications in our list.\label{tab:ThemsAndApplications}}
\end{table*}

\end{document}